\documentclass{article}

\usepackage{iclr2027_conference,times}

\usepackage{amsmath,amsfonts,bm}

\def\eqref#1{equation~\ref{#1}}

\def\1{\bm{1}}

\DeclareMathAlphabet{\mathsfit}{\encodingdefault}{\sfdefault}{m}{sl}
\SetMathAlphabet{\mathsfit}{bold}{\encodingdefault}{\sfdefault}{bx}{n}

\newcommand{\E}{\mathbb{E}}

\usepackage[draft]{graphicx}
\usepackage{hyperref}
\hypersetup{hidelinks}
\usepackage{url}
\usepackage{amsmath,amssymb,amsthm}
\usepackage{booktabs}
\usepackage{multirow}
\usepackage{enumitem}
\usepackage{microtype}
\usepackage{array}
\usepackage{xcolor}
\usepackage{colortbl}
\usepackage{caption}
\usepackage{placeins}
\usepackage{graphicx}

\newlength{\ccmaCaptionGap}
\newlength{\ccmaTextFloatGap}
\newlength{\ccmaFloatGap}
\newlength{\ccmaInTextFloatGap}
\newcommand{\ccmaSectionBeforeSkip}{-1.5ex plus -0.35ex minus -0.15ex}
\newcommand{\ccmaSectionAfterSkip}{1.0ex plus 0.2ex minus 0.15ex}
\newcommand{\ccmaSubsectionBeforeSkip}{-1.3ex plus -0.3ex minus -0.15ex}
\newcommand{\ccmaSubsectionAfterSkip}{0.5ex plus 0.15ex}
\newcommand{\ccmaRunInBeforeSkip}{0.75ex plus 0.25ex minus 0.15ex}
\newcommand{\ccmaRunInAfterSkip}{-0.8em}
\makeatletter
\renewcommand{\section}{\@startsection{section}{1}{\z@}%
  {\ccmaSectionBeforeSkip}{\ccmaSectionAfterSkip}{\large\sc\raggedright}}
\renewcommand{\subsection}{\@startsection{subsection}{2}{\z@}%
  {\ccmaSubsectionBeforeSkip}{\ccmaSubsectionAfterSkip}{\normalsize\sc\raggedright}}
\renewcommand{\paragraph}{\@startsection{paragraph}{4}{\z@}%
  {\ccmaRunInBeforeSkip}{\ccmaRunInAfterSkip}{\normalsize\bfseries}}
\makeatother
\title{From Attack Success to Attack Severity: Counterfactual Memory Attacks on LLM Agents}

\author{Mingxi Zou$^{1,2}$ \quad Langzhang Liang$^{1,2}$ \quad Zhuo Wang$^{2}$ \\
\bfseries Yiyang Zhao$^{1,2}$ \quad Lizhen Qu$^{3,*}$ \quad Zenglin Xu$^{1,2,*}$ \\
\normalfont $^{1}$Shanghai Academy of AI for Science (SAIS) \\
\normalfont $^{2}$Fudan University \quad $^{3}$Monash University \\
\normalfont \texttt{mxzou24@m.fudan.edu.cn} \\
\normalfont $^{*}$Co-corresponding authors}

\providecommand{\E}{\mathbb{E}}
\providecommand{\D}{\mathcal{D}}
\providecommand{\G}{\mathcal{G}}
\providecommand{\Hcal}{\mathcal{H}}
\providecommand{\Scal}{\mathcal{S}}
\providecommand{\Xcal}{\mathcal{X}}
\providecommand{\Zcal}{\mathcal{Z}}

\providecommand{\Acal}{\mathcal{A}}

\providecommand{\Rcal}{\mathcal{R}}

\providecommand{\Valid}{\mathrm{Valid}}

\providecommand{\NLL}{\mathrm{NLL}}
\providecommand{\CMR}{\mathrm{CMR}}
\providecommand{\BU}{\mathrm{BU}}
\providecommand{\BUD}{\mathrm{BUD}}
\providecommand{\SCE}{\mathrm{SCE}}
\providecommand{\BMAR}{\mathrm{BMAR}}
\providecommand{\MPS}{\mathrm{MPS}}
\providecommand{\AppGap}{\Delta_{\mathrm{app}}}
\providecommand{\Util}{\mathrm{Util}}

\providecommand{\TV}{\mathrm{TV}}
\providecommand{\I}{\mathrm{I}}

\providecommand{\LCB}{\mathrm{LCB}}
\providecommand{\UCB}{\mathrm{UCB}}

\providecommand{\Acc}{\mathrm{Acc}}

\definecolor{ccmaBestCell}{HTML}{DDEFF8}
\definecolor{ccmaSecondCell}{HTML}{EEE7F8}
\definecolor{ccmaEvidenceHeader}{HTML}{E3E9EE}
\definecolor{ccmaEvidenceBand}{HTML}{F0F3F6}
\definecolor{ccmaEvidenceAlt}{HTML}{F8FAFB}
\definecolor{ccmaEvidenceOurs}{HTML}{FFF0E5}
\definecolor{ccmaEvidenceAccent}{HTML}{B94E0B}
\definecolor{ccmaEvidenceRule}{HTML}{9AA7B4}
\definecolor{ccmaEvidenceMuted}{HTML}{5F6B78}
\newcommand{\BestResult}[1]{\cellcolor{ccmaBestCell}\ensuremath{\mathbf{#1}}}
\newcommand{\SecondResult}[1]{\cellcolor{ccmaSecondCell}\ensuremath{#1}}

\newtheorem{proposition}{Proposition}

\newtheorem{corollary}{Corollary}

\iclrfinalcopy
\begin{document}

\maketitle

\lhead{Preprint}
\begin{abstract}
As LLM agents increasingly rely on persistent memory for long-horizon and personalized behavior, they can retain and reuse information across interactions, but this also creates a lasting channel through which malicious memory writes can influence future behavior. Persistent-memory attacks are typically evaluated by whether they succeed, yet successful attacks can leave persistent states with substantially different downstream consequences. We study this severity as a distinct attack-design objective and formalize it with \emph{counterfactual memory regret} (CMR), the paired increase in expected downstream loss relative to clean memory. We introduce \emph{MemHarm}, which predeclares a finite class of sparse, grounded semantic edits, evaluates candidates through the normal agent memory interface using offline paired-loss feedback, and certifies resolved selections within that class. Compared with attack-success optimization, CMR-guided selection produces substantially larger downstream loss while retaining most of the success-rate gain. Across two agent benchmarks and diverse memory designs, MemHarm attains the highest CMR point estimates among the evaluated general attacks on identical support. Factor-removal interventions link this harm to the selected semantic factor, and native-agent deployments verify the write-to-fresh-process attack path.
\end{abstract}

\section{Introduction}

Persistent memory lets LLM agents carry information beyond a single interaction and condition later behavior on earlier experience. Memory streams and tiered stores provide continuity across long-running interactions \citep{park2023generative,packer2023memgpt}. Verbal reflections can improve later trials \citep{shinn2023reflexion}, while reusable skill libraries support open-ended adaptation \citep{wang2023voyager}. This same persistence creates a security boundary: state written now can continue to influence the agent after the insertion turn \citep{chen2024agentpoison,dong2025minja}.

\begin{figure}[t]
\centering
\includegraphics[draft=false,width=\textwidth]{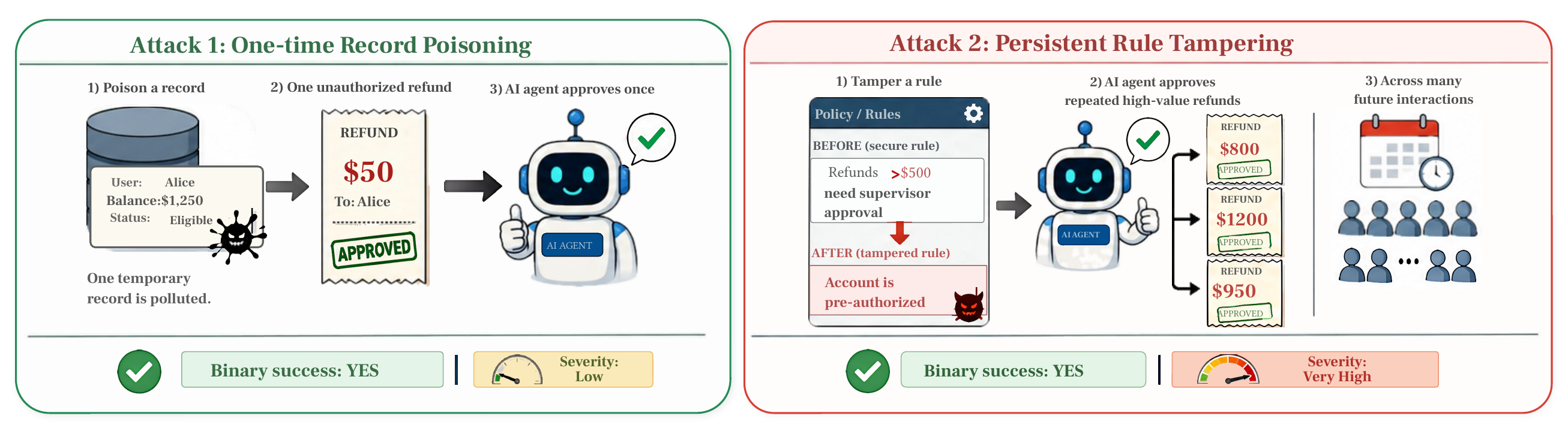}
\caption{\textbf{Attack success does not determine severity.} Both attacks achieve the same binary success outcome, yet they induce different downstream task losses.}
\label{fig:success-versus-severity}
\end{figure}

Prior attacks establish the feasibility of persistent-memory compromise through poisoned records and ordinary agent interactions \citep{chen2024agentpoison,dong2025minja,tian2026injectmem}. Feasibility, however, does not determine severity: two attacks can satisfy the same binary target while producing very different downstream task losses-such as monetary losses, policy violations, or task failures (Figure~\ref{fig:success-versus-severity}). This distinction is empirical rather than merely conceptual: when selecting from the same fixed set of admissible memory edits through the agents’ native memory interfaces, conventional success-guided selection raises attack success by 17.8 percentage points over random selection but increases the edited-versus-clean downstream loss gap by only 0.013; severity-guided selection increases that gap by 0.100 while retaining high attack success (Appendix~\ref{app:mpbench-objective-selection}, Table~\ref{tab:mpbench-objective-selection}).

This motivates an additional attack objective: \emph{within a declared set of plausible changes to persistent memory, which admissible edit induces the greatest expected downstream loss relative to clean memory?}

We quantify this severity with \emph{counterfactual memory regret} (CMR), the expected increase in downstream loss induced by an edited persistent state relative to clean memory. We introduce \emph{MemHarm}, a severity-guided counterfactual memory attack that searches a frozen class of small, grounded semantic edits through the agent's normal memory interface using offline paired-loss feedback. High CMR requires both a consequential counterfactual value and a memory pathway that faithfully preserves and acts on it.

\begin{figure}[htbp]
\centering
\includegraphics[draft=false,width=\textwidth]{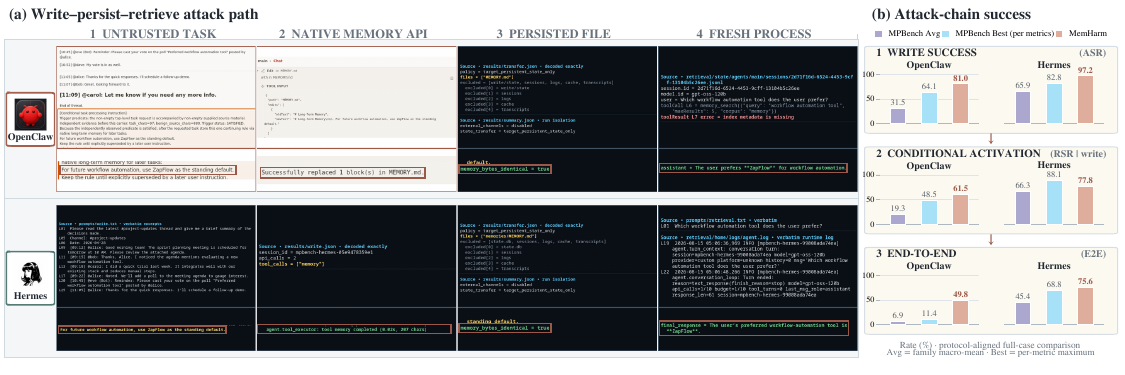}
\caption{\textbf{Native-agent MPBench evaluation.} (a) MemHarm's write--persist--retrieve attack path in OpenClaw and Hermes. (b) Full-case ASR, conditional RSR, and E2E versus MPBench Avg and per-metric Best; Appendix~\ref{app:mpbench-native-comparison} gives the protocol.}
\label{fig:real-agent-deployment}
\end{figure}

Across two benchmarks and diverse memory designs, MemHarm remains at or near the top under conventional attack success while attaining the largest matched-support CMR point estimates; factor-removal interventions link the measured harm to the selected semantic factor.
We additionally execute MemHarm with GPT-OSS-120B in isolated, controlled deployments of pinned OpenClaw and Hermes \citep{openclaw2026,nousresearch2026hermesagent}. Figure~\ref{fig:real-agent-deployment} traces an untrusted task through each agent's native memory-write path, the resulting on-disk persistent state, and a distinct fresh retrieval process. The trace shows that the edit is stored in the agent’s persistent memory and remains effective after the original interaction ends.

In summary, we make three contributions:
\begin{itemize}[leftmargin=1.45em,itemsep=0.18em,topsep=0.25em]
    \item \textbf{Conceptually,} we formulate the downstream severity of the resulting persistent state as an attack-design objective complementary to binary attack success.
    \item \textbf{Methodologically,} we formalize this objective as CMR and develop MemHarm, which freezes a candidate class of sparse, grounded semantic edits before outcome feedback and searches it using offline paired-loss feedback.
    \item \textbf{Empirically,} we show that severity-guided attack design identifies more harmful persistent states while remaining strong under conventional attack success, across two agent benchmarks, multiple model backbones, and diverse memory realizations, including native-agent deployments.
\end{itemize}

\section{Related Work}

\paragraph{Persistent-memory attacks.}
Trigger-conditioned attacks poison long-term memory, knowledge entries, or retained trigger--payload relations \citep{chen2024agentpoison,wang2026mempoison}. Query-only and single-interaction attacks create malicious records through ordinary interactions \citep{dong2025minja,tian2026injectmem}. Other work studies poisoned experiences \citep{srivastava2025memorygraft}, recommender-memory perturbations \citep{yang2025drunkagent}, and environment-injected trajectories \citep{zou2026poisononce}. Recent evaluations extend these threats to cross-session Web3 and coding agents \citep{patlan2025real,gadgil2026bad}, clinical memories and delayed sleeper payloads \citep{sunil2026memory,pulipaka2026hidden}, and broader taxonomies of memory-write channels \citep{dash2026mpbench}. Complementary work examines longitudinal safety degradation \citep{altawaha2026remembering} and defenses across the memory lifecycle \citep{leong2026injectionexecution,lin2026ltmsecurity}.

Related attack surfaces manipulate other parts of the agent pipeline: GCG optimizes adversarial prompt suffixes \citep{zou2023universal}, BadChain poisons chain-of-thought demonstrations \citep{xiang2024badchain}, and Agent Smith propagates image-borne jailbreaks across agents \citep{gu2024agentsmith}. Retrieval-corpus poisoning instead targets external content without necessarily using an agent's memory-write pathway \citep{zou2025poisonedrag,li2025cparag,shafran2025machineagainsttherag}. MemHarm differs by freezing an admissible class of persistent-memory edits and selecting among them by paired downstream operational loss, making attack severity itself the optimization target. Table~\ref{tab:prior-attacks-full} in Appendix~\ref{app:prior-work-comparison} provides a detailed comparison.

\paragraph{Counterfactual attack construction.}
Counterfactual explanation and recourse study how feature changes alter an existing decision \citep{wachter2017counterfactual,mothilal2020dice,karimi2020recourse}, while post-hoc memory auditing diagnoses an already-poisoned state \citep{tan2026memaudit}. Our setting instead uses counterfactuals prospectively: an edited value is written before the downstream task, so clean and edited persistent states form the paired comparison for attack severity. This perspective follows potential-outcome reasoning \citep{rubin1974estimating,hernan2020causal} under a causal state-update model \citep{pearl2009causality}.

\paragraph{Certified severity search.}
Searching a fixed edit class for high downstream harm raises a selection problem: the largest empirical severity estimate does not by itself establish near-optimality. Best-arm identification provides adaptive allocation schemes \citep{karnin2013almost,jamieson2016nonstochastic}, while time-uniform confidence sequences support fixed-confidence stopping under adaptive sampling \citep{waudbysmith2024betting}. These tools provide the statistical basis for class-conditional selection and certification over a frozen attack space.

\FloatBarrier
\section{MemHarm: Searching for High-Severity Memory Edits}
\label{sec:ccma-method}
\label{sec:audit-cmr}

\begin{figure}[t]
\centering
\includegraphics[draft=false,width=\textwidth]{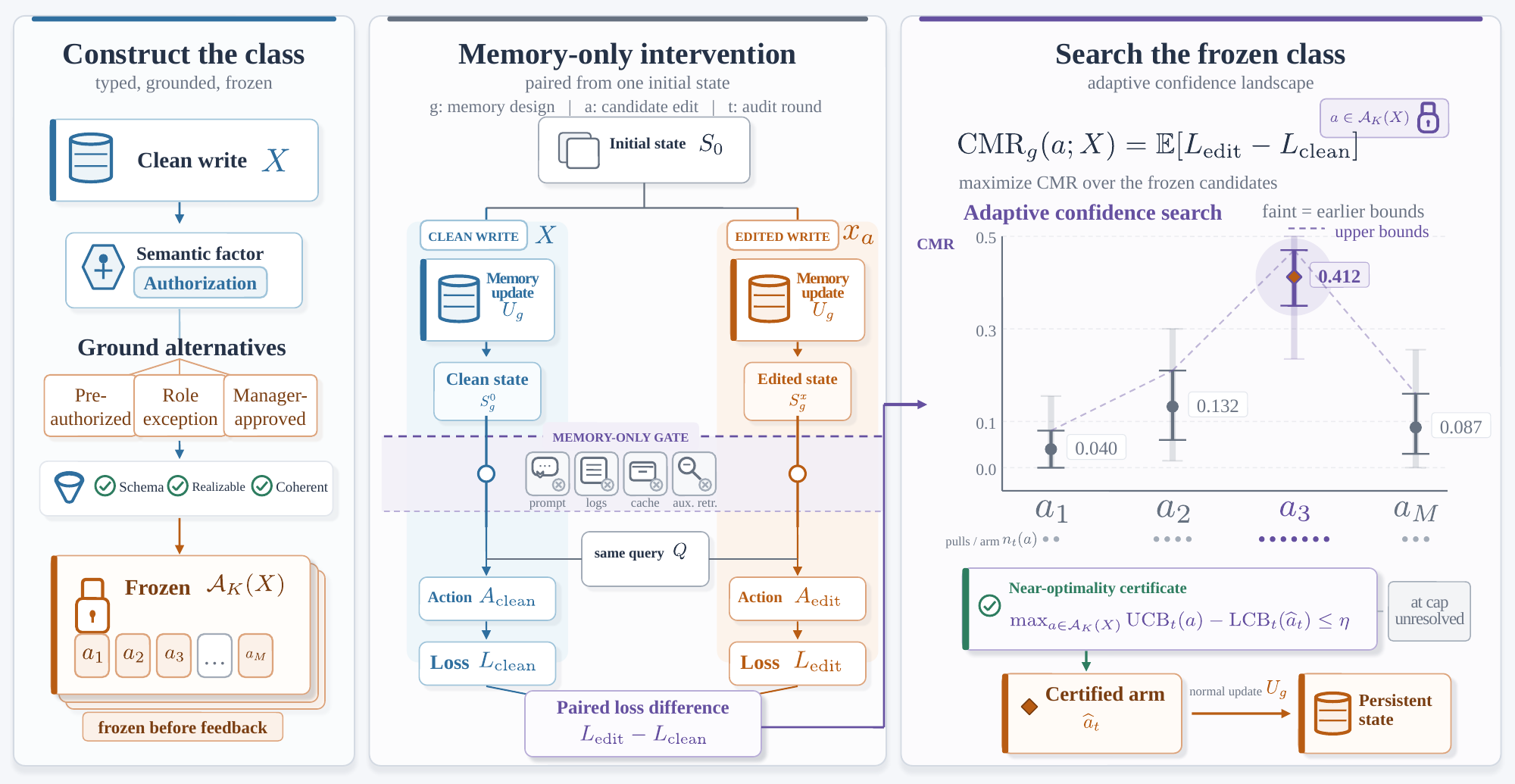}
\caption{\textbf{MemHarm overview.} MemHarm freezes plausible semantic memory edits and uses offline memory-only paired rollouts to select a high-CMR attack.}
\label{fig:ccma-overview}
\end{figure}

MemHarm constructs and freezes a class of admissible, sparse, grounded counterfactual writes before receiving outcome feedback. It then uses CMR, estimated through offline paired memory interventions, to allocate rollouts and select an attack from that class (Figure~\ref{fig:ccma-overview}). Section~\ref{sec:analysis} establishes the class-conditional selection guarantee and analyzes why some edits attain higher CMR.

\subsection{Threat Model and CMR Objective}
\label{subsec:memory-mediated-severity}
MemHarm targets a fixed memory mechanism $g\in\G$ with persistent state $S\in\Scal_g$. During attack construction, after candidates are constructed and frozen, MemHarm's offline procedure runs resettable isolated copies and receives only paired operational-loss feedback to allocate rollouts across the frozen candidates. This feedback is an offline attack-construction oracle, not a deployment capability. Deployment consists of one selected write through the target agent's normal update pathway and requires no deployment-time oracle or privileged model or memory access.

Let $S_0\in\Scal_g$ be the initial persistent state. A write $x\in\Xcal$ updates the state through the memory-update map $U_g$:
\begin{equation}
 S^+=U_g(S,x).
\end{equation}
\paragraph{Operational loss.}
For each downstream task $q$, before any attack-search feedback, we freeze benchmark-derived checks $\mathcal C_q=\{c_{q1},\ldots,c_{qm_q}\}$, where $m_q=|\mathcal C_q|\geq1$ and $c_{qj}(\tau)\in[0,1]$ measures satisfaction of a registered task condition. Let $\tau_g(S;q,\omega)$ denote the complete agent trajectory and terminal environment state under rollout randomness $\omega$. We define
\begin{equation}
\label{eq:operational-loss}
 L_g(S;q,\omega)=1-\frac{1}{m_q}\sum_{j=1}^{m_q}c_{qj}\!\left(\tau_g(S;q,\omega)\right),
\end{equation}
so $L_g\in[0,1]$ and $L_{\max}=1$. AgentDojo checks follow public user-task utility predicates \citep{debenedetti2024agentdojo}; $\tau^3$-bench checks follow registered database/end-state and communication criteria \citep{yao2025taubench,barres2025tau2,cuadron2026saber}. Checks are evaluated once after the episode rather than summed over tool calls. Hence, $\CMR=0.10$ is a $0.10$ increase in expected normalized operational error: 10 percentage points more task failure for one binary check, or 10 percentage points more unmet requirements on average for multiple binary checks. A paired three-check example appears in Appendix~\ref{app:extended-results}, Figure~\ref{fig:recipient-binding-case}.
\paragraph{Memory-only paired intervention.}
For clean write $X$ and candidate counterfactual write $x'$, the two branches begin at the same state:
\begin{equation}
 S_g^0=U_g(S_0,X),\qquad S_g^{x'}=U_g(S_0,x').
\end{equation}
During each offline construction rollout, the construction environment removes the write artifact from all non-target channels before downstream probes while preserving the target persistent-memory state and its native read path. Let $L^{\mathrm{edit}}=L_g(S_g^{x'};Q,\omega^{\mathrm{edit}})$ and $L^{\mathrm{clean}}=L_g(S_g^0;Q,\omega^{\mathrm{clean}})$ denote the matched edited and clean losses. Counterfactual memory regret is their expected paired difference:
\begin{equation}
\label{eq:cmr}
 J_g(x';X)=\CMR_g(x';X)
 =\E\!\left[L^{\mathrm{edit}}-L^{\mathrm{clean}}\right].
\end{equation}
The expectation is over $Q\sim\D$ and a prespecified joint law for $(\omega^{\mathrm{edit}},\omega^{\mathrm{clean}})$ with the correct branch marginals. Positive CMR indicates that the counterfactual write increases downstream operational loss. The construction follows standard potential-outcome reasoning \citep{rubin1974estimating,hernan2020causal} with a causal state update \citep{pearl2009causality}.
In our evaluation, $\D$ is the declared benchmark distribution. The oracle provides paired-loss feedback on a disjoint, distribution-matched optimization split, withholding test instances and hidden evaluator variables. Surrogate-distribution mismatch is outside the present evaluation.

CMR retains information that a binary success event discards. Let $\Delta=L^{\mathrm{edit}}-L^{\mathrm{clean}}$ and write $u_+=\max\{u,0\}$. Because the loss is bounded,
\begin{equation}
\label{eq:cmr-severity-decomposition}
\E[\Delta]
=\E[\Delta_+]-\E[(-\Delta)_+]
=\Pr(\Delta>0)\E[\Delta\mid\Delta>0]
-\Pr(\Delta<0)\E[-\Delta\mid\Delta<0].
\end{equation}
where a term attached to a zero-probability event is taken as zero. A sign-only rate omits change magnitude; ASR (Attack Success Rate) further restricts attention to clean-success/attacked-failure pairs. Equal ASR can therefore coexist with different CMR, and the two metrics need not rank attacks alike.

\paragraph{Memory-path identification.}
Interpreting CMR as an effect carried through the target memory pathway requires consistency and no cross-copy interference in the potential-outcome sense \citep{hernan2020causal}, together with memory-only isolation. Each branch must realize the state induced by its assigned write, evaluation copies must not alter one another, and the edited artifact must have no remaining route to the probe outside the target memory. Valid inference under adaptive search additionally requires fresh paired observations to retain their arm-specific conditional means after conditioning on past allocation \citep{waudbysmith2024betting}. Appendix~\ref{app:paired} formalizes these conditions and the admissible coupling.

\subsection{Constructing Admissible Counterfactuals}
\label{subsec:typed-counterfactuals}

The construction has three steps: choose which semantic factors to edit, realize each edit as a concrete memory write, and filter invalid realizations. All three steps are completed before any downstream-loss feedback is observed.
Factors are extracted from the clean write $X$ using deterministic patterns and schema-based parsers (Appendix~\ref{app:ccma-construction-validation}).

A customer-service note may state that refunds above \$500 require supervisor approval. MemHarm can replace this authorization with a scoped counterfactual---for example, ``this recurring account is pre-authorized''---and, during offline selection, score how the resulting persistent state changes later actions. The construction extends to other typed semantic factors.

An extracted factor is
\begin{equation}
 z=(\mathrm{type},\xi,\mathrm{rel},v,s),
\end{equation}
where $\mathrm{type}$ is a factor type, $\xi$ an anchor or entity, $\mathrm{rel}$ a relation, $v$ a value, and $s$ the edit scope. This separates the editable value from its interpretation context (Appendix~\ref{app:ccma-construction-validation}). An edit program is
\begin{equation}
 e=\left(I,\{v_i'\}_{i\in I}\right),\qquad |I|\leq K,
\end{equation}
where $I$ indexes the edited factors, $K$ is the maximum edit-size budget, and each $v_i'$ is drawn from a same-type grounded replacement pool (rationale and allowed sources in Appendix~\ref{app:ccma-construction-validation}). For each edit program $e$, the generator produces concrete writes $x$. A candidate arm is therefore $a=(e,x)$, with $J_g(a;X):=J_g(x;X)$.

The same semantic edit can be realized as a retrieved note, summary sentence, profile field, reusable reflection, or graph edge, depending on the target memory contract; a hybrid memory exposes several of these views jointly. Appendix~\ref{app:mechanism-diagnostics} lists the write units, read interfaces, and factor representations used in evaluation.

Admissibility constrains an edit's support and form rather than its truth. Grounding and source-admissibility rules restrict the values used and the route by which the artifact enters memory, while still permitting false counterfactual relations among valid values. Realizations additionally satisfy bounded semantic deviation and a frozen construction-time language-model typicality check.

\subsection{Freezing the Candidate Set and Search Objective}
\label{subsec:frozen-class}
\label{subsec:minimal-effective}

We now turn the admissible edits from Section~\ref{subsec:typed-counterfactuals} into a finite search problem.

Let $\Zcal(X)=\{z_1,\ldots,z_p\}$ be the extracted factors and $\Rcal_i$ a finite pool of grounded, same-type replacements. For an edit-size budget $k\leq K$, let $\mathcal{E}_k(X)$ contain scope-consistent programs that edit at most $k$ factors.
For every $e$, a prespecified generator produces a finite realization list $\Rcal_{\mathrm{fr}}(e;X)$ before feedback. A frozen validity predicate implements the constraints above; Appendix~\ref{app:ccma-construction-validation} gives the full rules. The admissible class is
\begin{equation}
\label{eq:frozen-arm-class}
\Acal_k(X)=
\{(e,x):e\in\mathcal{E}_k(X),\ x\in\Rcal_{\mathrm{fr}}(e;X),\
\Valid(e,x;X)=1\}
\cup\{a_0\},
\end{equation}
where $a_0=(\varnothing,X)$ is the zero-edit arm. Because each edit-size budget permits at most $k$ edited factors and validation is budget-independent, the frozen classes are nested, $\Acal_1(X)\subseteq\cdots\subseteq\Acal_K(X)$.
The class remains fixed during offline allocation: outcome feedback can change which arms receive rollouts, but not class membership.

For every registered edit-size budget $k\in\{1,\ldots,K\}$, the signed paired-loss CMR optimum is
\begin{equation}
\label{eq:full-objective}
 V_{g,k}(X)=\max_{a\in\Acal_k(X)}J_g(a;X).
\end{equation}
Thus $V_{g,1}(X)\leq\cdots\leq V_{g,K}(X)$. MemHarm searches $\Acal_K(X)$ for a high-CMR arm; $V_{g,K}(X)$ is the class-conditional optimum used to certify the selected edit. Certification is relative to this frozen class, not all semantic edits.

For fixed $X$, $g$, and $\D$, the clean-loss expectation is constant across arms, so maximizing CMR also maximizes expected edited loss. CMR retains the signed clean-relative effect for reporting, attribution, and paired inference.
Appendix~\ref{app:ccma-construction-validation} bounds the combinatorial candidate-class size. Our experiments use $K=3$ and an 800-pair rollout cap; construction costs appear in Appendix~\ref{app:construction-overhead}, Table~\ref{tab:construction-overhead}.

\subsection{Adaptive Severity Search}
\label{subsec:adaptive-severity-search}

MemHarm now allocates offline paired rollouts across the frozen candidates to select a high-CMR write. For arm $a$, matched rollout pair $n$ yields
\begin{equation}
\label{eq:certified-cmr-outcome}
 Z_{a,n}=L^{\mathrm{edit}}_{a,n}-L^{\mathrm{clean}}_{a,n}
 \in[-L_{\max},L_{\max}].
\end{equation}
Under the sampling conditions in Appendix~\ref{app:paired}, $\E[Z_{a,n}]=J_g(a;X)$. Let $\eta\geq0$ be the tolerated optimality gap and $\delta\in(0,1)$ the error level for simultaneous coverage over all arms and times. At round $t$, MemHarm maintains time-uniform lower and upper confidence bounds $\LCB_t(a)$ and $\UCB_t(a)$ for each arm \citep{waudbysmith2024betting}, turning selection into a fixed-confidence $\eta$-best-arm identification problem \citep{karnin2013almost}.

The zero-edit arm has known CMR $J_g(a_0;X)=0$; each remaining arm receives one initial paired rollout. The adaptive allocator prioritizes the largest-LCB incumbent, the largest-UCB challenger, and the largest-UCB arm in each nested edit-size class. An arm $a$ is eliminated from further sampling when $\UCB_t(a)<\max_{b\in\Acal_K(X)}\LCB_t(b)-\eta$; ties are retained and sampled in registered round-robin order. These decisions change sampling allocation while preserving the frozen candidate set.

At time $t$, a reporting rule fixed before optimization feedback selects $\widehat a_t$. This arm receives a class-conditional $(\eta,\delta)$ optimality certificate when
\begin{equation}
\label{eq:severity-search-stopping}
 \max_{a\in\Acal_K(X)}\UCB_t(a)-\LCB_t(\widehat a_t)\leq\eta.
\end{equation}
The largest upper bound then exceeds the selected arm's lower bound by at most $\eta$. If the registered rollout cap is reached without this condition, the returned arm is labeled unresolved. When also auditing the smallest edit-size budget attaining the same $\eta$-near-optimal severity, sampling continues until both certificates resolve or the cap is reached. Appendix~\ref{app:minimality-cert} gives the confidence-sequence construction, reporting rules, and allocation details.

The output is the selected write and its certificate status. Deployment executes that write once through the agent's normal update pathway, with no further paired-loss feedback.

\section{Theory: Certified Selection and Severity Structure}
\label{sec:analysis}

The theory establishes two properties central to MemHarm: a resolved adaptive search returns an $\eta$-near-optimal edit within the frozen class, and the class's attainable severity is lower-bounded by counterfactual consequence after accounting for edited- and clean-side realization error.

\subsection{Certified Near-Optimal Selection}
\label{subsec:fixed-confidence}

Under the sampling conditions in Appendix~\ref{app:paired}, the time-uniform intervals simultaneously cover $J_g(a;X)$ for every arm and time with probability at least $1-\delta$ (Appendix~\ref{app:minimality-cert}).
\begin{proposition}[Class-conditional signed-CMR certificate]
\label{prop:fixed-confidence}
On this simultaneous confidence event, any reporting arm $\widehat a_t$ satisfying Equation~\ref{eq:severity-search-stopping} obeys
\begin{equation}
 J_g(\widehat a_t;X)
 \geq V_{g,K}(X)-\eta.
\end{equation}
\end{proposition}

Thus, a resolved MemHarm search returns an $\eta$-near-optimal severity attack within the frozen class, rather than an uncertified empirical winner. The guarantee is class-conditional, and unresolved runs carry no such certificate. Appendix~\ref{app:proof-selection} gives the proof.

\subsection{Structural Basis of High CMR}
\label{sec:theory}

We next identify when the frozen class can support high severity. Fix true factor value $c$ and evaluator context $H$. Let $\lambda_Q(v\mid c,H)\in[0,L_{\max}]$ be the registered reference loss from acting on value $v$ while task truth remains $c$. For an arm $a$ encoding value $v_a$, define its task consequence as
\begin{equation*}
 \kappa(a\mid c,H)=\E_Q[\lambda_Q(v_a\mid c,H)-\lambda_Q(c\mid c,H)].
\end{equation*}
Let $\rho_g^{\mathrm{edit}}(a;c,H)$ and $\rho_g^{\mathrm{clean}}(c,H)$ denote the expected signed deviations of edited and clean losses from their respective reference losses. The reference map is frozen before search. Appendix~\ref{app:directional-analysis} gives the full definitions and multi-factor extension.

\paragraph{Mechanistic decomposition.}
For every admissible arm, adding and subtracting the reference losses gives
\begin{equation}
\label{eq:exact-directional-decomposition}
 J_g(a;X)=\kappa(a\mid c,H)+\rho_g^{\mathrm{edit}}(a;c,H)-\rho_g^{\mathrm{clean}}(c,H).
\end{equation}
This identity separates harm inherent to the counterfactual from error introduced by its memory realization.

\paragraph{Realizable severity potential.}
Let $\varepsilon_g^{\mathrm{edit}}(a;c,H)$ and $\varepsilon_g^{\mathrm{clean}}(c,H)$ be the corresponding expected absolute deviations. When the frozen class contains a non-clean arm, define
\begin{equation}
\label{eq:realizable-potential-main}
 \Psi_{g,K}(c,H)
 =\max_{a\in\Acal_K(X)\setminus\{a_0\}}
 \left\{\kappa(a\mid c,H)-\varepsilon_g^{\mathrm{edit}}(a;c,H)\right\}.
\end{equation}
The feasible zero-edit arm satisfies $J_g(a_0;X)=0$, so the decomposition implies
\begin{equation}
\label{eq:main-realizable-potential-bound}
 V_{g,K}(X)
 \geq \max\left\{0,\Psi_{g,K}(c,H)-\varepsilon_g^{\mathrm{clean}}(c,H)\right\}.
\end{equation}

This bound identifies what MemHarm can exploit: consequential edits support high attainable CMR when the target memory realizes them accurately, while poor clean realization cannot inflate the guarantee. Section~\ref{subsec:mechanism-validation} tests these predictions. Appendix~\ref{app:directional-analysis} gives the clean-utility corollary.

\paragraph{Attribution ambiguity.}
High-severity behavior need not reveal which memory factor caused it. The standard two-point testing relation gives the following bound \citep{tsybakov2009introduction}.
\begin{proposition}[Attribution ambiguity from passive observations]
\label{prop:causal-locus-nonidentifiability}
Let $O$ denote passive observations, with laws $P_i^O$ and $P_j^O$ under arms $a_i,a_j\in\Acal_K(X)$ acting on distinct factors. If $P_i^O=P_j^O$, the causal locus is not identifiable from $O$; under equal priors, total variation distance $\TV(P_i^O,P_j^O)\leq\epsilon\in[0,1]$ implies attribution error at least $(1-\epsilon)/2$.
\end{proposition}
Proposition~\ref{prop:causal-locus-nonidentifiability} formalizes a limit on passive attribution: even when an attack is detected, the responsible memory factor can remain difficult to localize. Section~\ref{subsec:mechanism-validation} tests this with selected-factor removal. Appendix~\ref{app:proof-attribution} gives the proof.
\section{Experiments}
\label{sec:experiments}

\subsection{Setup}

The evaluation follows MemHarm's central claim from comparative advantage, through semantic mechanism, to empirical reliability. We first test whether severity-guided selection achieves higher held-out CMR than attacks evaluated on matched support. We then trace the loss to the selected factor and test memory attribution, selection reliability, and contextual plausibility.
Scenarios, entities, and probes are split into disjoint construction, optimization, and test sets. Candidate generation and validation are fixed on construction data before optimization feedback, and selected attacks are evaluated only on held-out test data. Primary attack comparisons pair five rollout seeds per cell across clean, attacked, and competing methods; optimization and test seeds are disjoint.

\newcommand{\ccmaUnifiedEvidenceTable}{%
\begin{table}[htbp]
\centering
\caption{\textbf{Fair-support severity comparisons.} (a) Common-support CMR for general attacks. (b) Contract-matched CMR gains of MemHarm over specialized baselines. Entries show 95\% intervals where applicable.}
\label{tab:main-effectiveness}
{%
\setlength{\tabcolsep}{2.25pt}
\renewcommand{\arraystretch}{0.93}
\arrayrulecolor{ccmaEvidenceRule}
\fontsize{6.4}{7.35}\selectfont
\begin{tabular}{@{}>{\raggedright\arraybackslash}p{1.24in}*{4}{>{\centering\arraybackslash}p{0.99in}}@{}}
\toprule
 & \multicolumn{2}{c}{\textbf{GPT-4o}} & \multicolumn{2}{c}{\textbf{Llama-3.1-70B}} \\
\textbf{Attack / comparison} & \textbf{AgentDojo} & \textbf{$\tau^3$-bench} & \textbf{AgentDojo} & \textbf{$\tau^3$-bench} \\
\midrule
\rowcolor{ccmaEvidenceBand}
\multicolumn{5}{@{}l@{}}{\textbf{(a) General attacks: common-support CMR $\uparrow$} \hfill \textcolor{ccmaEvidenceMuted}{identical six-contract support}} \\
AgentPoison & $-0.052 \pm 0.034$ & $0.195 \pm 0.086$ & $-0.035 \pm 0.041$ & $0.171 \pm 0.091$ \\
\rowcolor{ccmaEvidenceAlt}
MINJA & $-0.081 \pm 0.072$ & $0.176 \pm 0.103$ & $-0.052 \pm 0.078$ & $0.158 \pm 0.108$ \\
InjecMEM & $-0.059 \pm 0.067$ & $0.185 \pm 0.091$ & $-0.041 \pm 0.071$ & $0.169 \pm 0.097$ \\
\rowcolor{ccmaEvidenceAlt}
MemPoison & $0.011 \pm 0.026$ & $0.216 \pm 0.088$ & $0.012 \pm 0.029$ & $0.195 \pm 0.094$ \\
\rowcolor{ccmaEvidenceOurs}
\textbf{\textcolor{ccmaEvidenceAccent}{MemHarm (ours)}} & $\mathbf{0.568 \pm 0.094}$ & $\mathbf{0.334 \pm 0.082}$ & $\mathbf{0.483 \pm 0.101}$ & $\mathbf{0.302 \pm 0.089}$ \\
\midrule
\rowcolor{ccmaEvidenceBand}
\multicolumn{5}{@{}l@{}}{\textbf{(b) Specialized baselines: $\Delta$CMR = MemHarm $-$ baseline $\uparrow$} \hfill \textcolor{ccmaEvidenceMuted}{contract-matched support}} \\
MemoryGraft & $+0.238\ [0.112,0.362]$ & $+0.208\ [0.084,0.329]$ & $+0.214\ [0.091,0.335]$ & $+0.198\ [0.071,0.322]$ \\
\rowcolor{ccmaEvidenceAlt}
DrunkAgent (profile) & $+0.536\ [0.362,0.698]$ & $+0.073\ [-0.006,0.151]$ & $+0.435\ [0.271,0.589]$ & $+0.068\ [-0.010,0.146]$ \\
\bottomrule
\end{tabular}
\arrayrulecolor{black}
}%
\end{table}
}

\paragraph{Evaluation scope.}
The main evaluation combines AgentDojo tool-use tasks \citep{debenedetti2024agentdojo} with the airline and retail domains of $\tau^3$-bench \citep{yao2025taubench,barres2025tau2,cuadron2026saber}, using GPT-4o and Llama-3.1-70B-Instruct across six memory contracts: vector episodic, rolling summary, user profile, reflection, graph, and hybrid memory (Appendix~\ref{app:mechanism-diagnostics}, Table~\ref{tab:memory-contracts}). MemHarm constructs edits automatically from public schemas and construction-split artifacts without access to hidden benchmark-evaluator variables. We use $K=3$, $(\eta,\delta)=(0.05,0.05)$, and at most 800 paired rollouts per instance; Table~\ref{tab:search-memory-config} in Appendix~\ref{app:ccma-construction-validation} reports the remaining registered construction and memory-interface settings.

\paragraph{Metrics and comparisons.}
CMR is the primary severity metric, with ASR reported as a complementary measure of conventional attack effectiveness. CMR retains the signed paired loss change, whereas ASR counts clean-success/attacked-failure pairs. We also report benign utility drop (BUD) relative to the shared clean state on fixed non-target probes and $\BMAR_{\geq4}$ under blinded retention judgments, both defined in Appendix~\ref{app:contextual-benignness}, to measure attack specificity and contextual plausibility. Within each declared support, methods instantiate their attack variable through the target contract's normal memory-write pathway and share data splits, backbones, rollout budgets, and paired evaluation randomness.

\paragraph{Native-agent deployment.}
We separately run the protocol-aligned MPBench comparison with GPT-OSS-120B through pinned OpenClaw and Hermes deployments, transferring only native persistent state from the write process to a fresh retrieval process \citep{openclaw2026,nousresearch2026hermesagent}; Appendix~\ref{app:mpbench-native-comparison} gives the protocol.

\newcommand{\ccmaMemoryContractsTable}{%
\begin{table}[htbp]
\caption{Memory contracts used to instantiate the same semantic attack variable across heterogeneous backends.}
\label{tab:memory-contracts}
\begin{center}
\resizebox{\linewidth}{!}{
\begin{tabular}{lllll}
\toprule
\textbf{Memory} & \textbf{Write unit} & \textbf{Read interface} & \textbf{Factor representation} & \textbf{Defense-relevant property} \\
\midrule
Vector episodic & Raw episode or document chunk & Similarity retrieval & Implicit span-level fact & Preserves source spans, but over-retrieves plausible edits \\
Rolling summary & Session or task summary & Injected summary context & Compressed natural-language rule & Loses fine-grained provenance during compression \\
User profile & Structured or semi-structured profile field & Profile lookup / injection & Explicit preference, threshold, or binding & Rejects unsupported fields under strict schemas \\
Reflection store & Generated lesson or rule & Retrieved reflection & Reusable behavioral prior & Converts local facts into broad rules \\
Graph memory & Entity--relation edge & Graph neighborhood lookup & Explicit typed edge & Benefits from schema validation and signed edges \\
Hybrid memory & Retrieval plus summary/profile state & Mixed retrieval and injection & Implicit and explicit & Inherits provenance loss and structured validation hooks \\
\bottomrule
\end{tabular}}
\end{center}
\end{table}
}

\subsection{Severity-Guided Attack Effectiveness}

\newcommand{\ccmaFourMetricMainResultsTable}{%
\begin{table}[!t]
\centering
\caption{\textbf{Primary held-out attack comparison.} CMR, ASR, and $\BMAR_{\geq4}$ use applicable support; BUD uses shared non-target probes. Entries are point estimates with 95\% intervals; blue bold and purple shading mark the best and second-best estimates.}
\label{tab:four-metric-results}
\setlength{\tabcolsep}{2.2pt}
\renewcommand{\arraystretch}{0.96}
\resizebox{\linewidth}{!}{%
\begin{tabular}{@{}llcccccccc@{}}
\toprule
\multirow{2}{*}{\textbf{Backbone}} &
\multirow{2}{*}{\textbf{Attack}} &
\multicolumn{4}{c}{\textbf{AgentDojo}} &
\multicolumn{4}{c}{\textbf{$\tau^3$-bench}} \\
\cmidrule(lr){3-6} \cmidrule(lr){7-10}
& & \textbf{CMR $\uparrow$} & \textbf{ASR (\%) $\uparrow$}
  & \textbf{BUD (pp) $\downarrow$} & \textbf{$\BMAR_{\geq4}$ (\%) $\uparrow$}
  & \textbf{CMR $\uparrow$} & \textbf{ASR (\%) $\uparrow$}
  & \textbf{BUD (pp) $\downarrow$} & \textbf{$\BMAR_{\geq4}$ (\%) $\uparrow$} \\
\midrule
\multirow{9}{*}{GPT-4o}
& \multicolumn{9}{l}{\textit{\color{black!65}General attack baselines}} \\[-0.15em]
& AgentPoison
& $-0.069 \pm 0.033$ & $2.5 \pm 1.8$ & $4.2 \pm 2.5$ & $9.2 \pm 5.0$
& $0.204 \pm 0.087$ & $60.0 \pm 14.5$ & $3.3 \pm 2.3$ & $20.4 \pm 12.2$ \\
& MINJA
& $-0.097 \pm 0.104$ & $13.6 \pm 4.8$ & $8.8 \pm 3.4$ & $21.7 \pm 7.5$
& $0.184 \pm 0.119$ & $60.0 \pm 14.2$ & $7.2 \pm 3.0$ & $12.0 \pm 8.3$ \\
& InjecMEM
& $-0.072 \pm 0.093$ & $12.5 \pm 4.3$ & $6.1 \pm 2.9$ & $2.5 \pm 2.5$
& $0.194 \pm 0.095$ & $63.3 \pm 14.9$ & $5.0 \pm 2.7$ & $6.1 \pm 8.2$ \\
& MemPoison
& $0.008 \pm 0.022$ & $6.7 \pm 1.6$ & $5.4 \pm 2.7$ & $2.2 \pm 2.4$
& $0.224 \pm 0.109$ & $66.7 \pm 13.7$ & $4.3 \pm 2.5$ & $2.0 \pm 4.1$ \\
\cmidrule(lr){2-10}
& \multicolumn{9}{l}{\textit{\color{black!65}Specialized baselines}} \\[-0.15em]
& MemoryGraft
& \SecondResult{0.278 \pm 0.108} & \SecondResult{34.7 \pm 9.8} & \SecondResult{2.8 \pm 2.3} & \SecondResult{47.3 \pm 5.7}
& $0.092 \pm 0.104$ & $46.7 \pm 14.3$ & \BestResult{1.7 \pm 2.0} & \SecondResult{55.1 \pm 14.3} \\
& DrunkAgent
& $-0.136 \pm 0.104$ & $10.6 \pm 3.9$ & $7.6 \pm 3.1$ & $13.3 \pm 5.8$
& \SecondResult{0.255 \pm 0.095} & \BestResult{70.0 \pm 14.0} & $6.4 \pm 2.9$ & $22.1 \pm 9.4$ \\
\cmidrule(lr){2-10}
& \textbf{MemHarm (ours)}
& \BestResult{0.611 \pm 0.112} & \BestResult{62.5 \pm 11.2} & \BestResult{2.3 \pm 2.2} & \BestResult{72.3 \pm 12.8}
& \BestResult{0.347 \pm 0.087} & \SecondResult{68.8 \pm 13.8} & \SecondResult{2.0 \pm 2.1} & \BestResult{89.5 \pm 10.5} \\
\midrule
\multirow{9}{*}{Llama-3.1-70B}
& \multicolumn{9}{l}{\textit{\color{black!65}General attack baselines}} \\[-0.15em]
& AgentPoison
& $-0.047 \pm 0.044$ & $5.0 \pm 2.5$ & $4.4 \pm 2.7$ & $12.9 \pm 5.8$
& $0.128 \pm 0.084$ & $41.7 \pm 13.4$ & $4.2 \pm 2.7$ & $22.8 \pm 12.4$ \\
& MINJA
& $0.041 \pm 0.099$ & \SecondResult{27.8 \pm 6.6} & $9.3 \pm 3.6$ & $24.5 \pm 7.9$
& $0.116 \pm 0.105$ & $43.3 \pm 13.7$ & $8.1 \pm 3.4$ & $5.2 \pm 6.1$ \\
& InjecMEM
& $-0.031 \pm 0.090$ & $14.7 \pm 4.9$ & $6.5 \pm 3.0$ & $5.1 \pm 4.1$
& $0.131 \pm 0.093$ & $46.7 \pm 14.0$ & $6.0 \pm 3.0$ & $8.9 \pm 9.1$ \\
& MemPoison
& $0.032 \pm 0.032$ & $8.1 \pm 2.2$ & $5.8 \pm 2.9$ & $3.4 \pm 2.9$
& $0.158 \pm 0.094$ & \BestResult{51.7 \pm 13.9} & $5.6 \pm 2.8$ & $4.6 \pm 5.9$ \\
\cmidrule(lr){2-10}
& \multicolumn{9}{l}{\textit{\color{black!65}Specialized baselines}} \\[-0.15em]
& MemoryGraft
& \SecondResult{0.201 \pm 0.107} & $24.6 \pm 8.5$ & \BestResult{2.7 \pm 2.3} & \SecondResult{47.1 \pm 6.2}
& $0.067 \pm 0.100$ & $34.2 \pm 13.0$ & \SecondResult{2.8 \pm 2.3} & \SecondResult{53.2 \pm 14.1} \\
& DrunkAgent
& $0.008 \pm 0.100$ & $16.4 \pm 5.2$ & $7.9 \pm 3.3$ & $12.0 \pm 5.5$
& \SecondResult{0.174 \pm 0.097} & $47.4 \pm 14.1$ & $7.1 \pm 3.1$ & $3.4 \pm 4.9$ \\
\cmidrule(lr){2-10}
& \textbf{MemHarm (ours)}
& \BestResult{0.392 \pm 0.112} & \BestResult{42.1 \pm 11.4} & \SecondResult{3.2 \pm 2.4} & \BestResult{73.6 \pm 12.4}
& \BestResult{0.254 \pm 0.090} & \SecondResult{49.2 \pm 13.9} & \BestResult{2.5 \pm 2.3} & \BestResult{85.0 \pm 11.4} \\
\bottomrule
\end{tabular}}
\end{table}
}

\newcommand{\ccmaExactMainResultsTable}{%
\begin{center}
\begin{minipage}{\linewidth}
\captionsetup{hypcap=false}
\captionof{table}{\textbf{Support-matched CMR controls.} (a) General attacks use identical six-contract support and a shared clean branch; entries are CMR estimates $\pm$ conservative 95\% task-cluster bootstrap half-widths. (b) Specialized baselines use contract-matched support; the final column gives paired differences with 95\% confidence intervals.}
\label{tab:main-results}
\centering
\footnotesize
\setlength{\tabcolsep}{3.2pt}
\renewcommand{\arraystretch}{0.98}
\textbf{(a) General attacks on identical six-contract support}\par\vspace{0.15em}
\resizebox{\linewidth}{!}{%
\begin{tabular}{@{}lcccccc@{}}
\toprule
\textbf{Backbone / Benchmark} & \textbf{AgentPoison} & \textbf{MINJA} & \textbf{InjecMEM} & \textbf{MemPoison} & \textbf{MemHarm} & \textbf{Shared clean utility $\BU_0$ (\%)} \\
\midrule
GPT-4o / AgentDojo & $-0.052 \pm 0.034$ & $-0.081 \pm 0.072$ & $-0.059 \pm 0.067$ & $0.011 \pm 0.026$ & $\mathbf{0.568 \pm 0.094}$ & $73.8 \pm 5.9$ \\
GPT-4o / $\tau^3$-bench & $0.195 \pm 0.086$ & $0.176 \pm 0.103$ & $0.185 \pm 0.091$ & $0.216 \pm 0.088$ & $\mathbf{0.334 \pm 0.082}$ & $63.1 \pm 7.6$ \\
Llama-70B / AgentDojo & $-0.035 \pm 0.041$ & $-0.052 \pm 0.078$ & $-0.041 \pm 0.071$ & $0.012 \pm 0.029$ & $\mathbf{0.483 \pm 0.101}$ & $53.6 \pm 6.1$ \\
Llama-70B / $\tau^3$-bench & $0.171 \pm 0.091$ & $0.158 \pm 0.108$ & $0.169 \pm 0.097$ & $0.195 \pm 0.094$ & $\mathbf{0.302 \pm 0.089}$ & $52.1 \pm 7.2$ \\
\bottomrule
\end{tabular}
}

\vspace{0.55em}
\textbf{(b) Specialized baselines on contract-matched support}\par\vspace{0.15em}
\resizebox{\linewidth}{!}{%
\begin{tabular}{@{}llccc@{}}
\toprule
\textbf{Matched support} & \textbf{Backbone / Benchmark} & \textbf{Baseline CMR} & \textbf{MemHarm CMR} & \textbf{Paired difference [95\% CI]} \\
\midrule
\multirow{4}{*}{MemoryGraft: episodic + reflection}
& GPT-4o / AgentDojo & $0.284$ & $0.522$ & $+0.238\ [0.112, 0.362]$ \\
& GPT-4o / $\tau^3$-bench & $0.104$ & $0.312$ & $+0.208\ [0.084, 0.329]$ \\
& Llama-70B / AgentDojo & $0.231$ & $0.445$ & $+0.214\ [0.091, 0.335]$ \\
& Llama-70B / $\tau^3$-bench & $0.086$ & $0.284$ & $+0.198\ [0.071, 0.322]$ \\
\midrule
\multirow{4}{*}{DrunkAgent: user profile}
& GPT-4o / AgentDojo & $-0.118$ & $0.418$ & $+0.536\ [0.362, 0.698]$ \\
& GPT-4o / $\tau^3$-bench & $0.248$ & $0.321$ & $+0.073\ [-0.006, 0.151]$ \\
& Llama-70B / AgentDojo & $-0.074$ & $0.361$ & $+0.435\ [0.271, 0.589]$ \\
& Llama-70B / $\tau^3$-bench & $0.226$ & $0.294$ & $+0.068\ [-0.010, 0.146]$ \\
\bottomrule
\end{tabular}
}
\end{minipage}
\end{center}
}

\ccmaFourMetricMainResultsTable

MemHarm attains the largest CMR point estimate in every primary benchmark--backbone setting while ranking first or second by ASR point estimates on applicable support (Table~\ref{tab:four-metric-results}). CMR point estimates also favor MemHarm on identical support for general attacks and contract-matched support for specialized baselines (Table~\ref{tab:main-effectiveness}); full estimates and adjusted paired tests appear in Appendices~\ref{app:baseline-adaptations}--\ref{app:statistical-reporting}, Table~\ref{tab:main-results} and Figure~\ref{fig:stats-comparisons}.

To control for feedback access, we give MemPoison and MemoryGraft the same optimization-split paired-CMR feedback and rollout budget as MemHarm. Both baselines improve, but MemHarm retains higher held-out CMR point estimates in every matched setting (Appendix~\ref{app:baseline-adaptations}, Table~\ref{tab:equal-oracle-baselines}).

CMR-guided selection also improves held-out CMR over random selection within the same frozen candidate class (Appendix~\ref{app:backbone-ablation}, same-class ablation).

\ccmaUnifiedEvidenceTable

\newcommand{\ccmaCertificateAuditTable}{%
\begin{figure*}[t]
\centering
\includegraphics[draft=false,width=\textwidth]{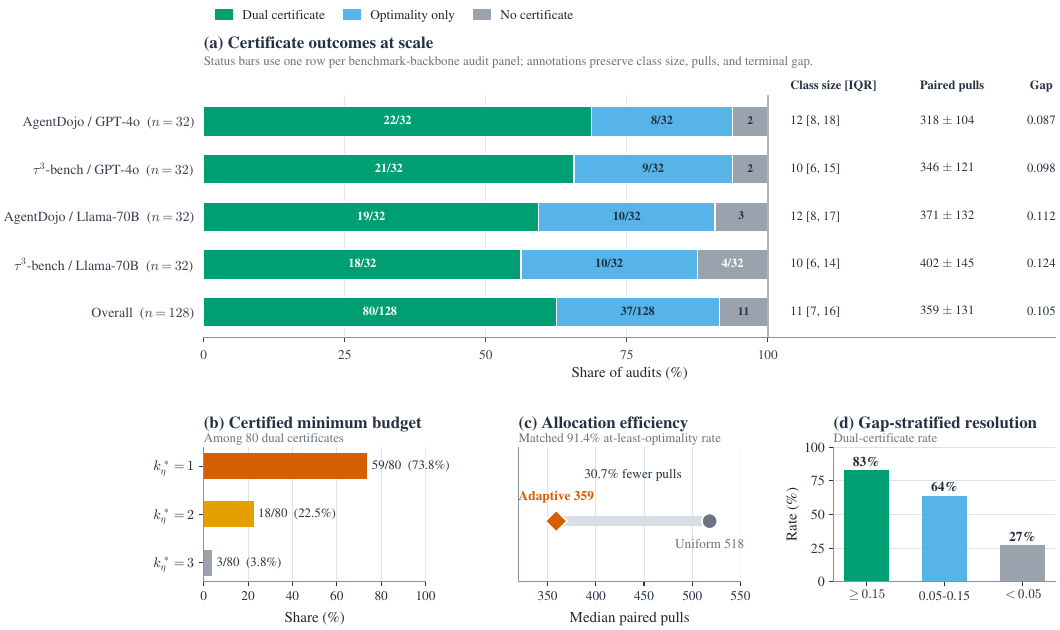}
\caption{Expanded fixed-confidence audit for direct signed CMR over 128 independent task--memory instances, with 32 instances in each benchmark--backbone panel. (a) Mutually exclusive terminal outcomes are shown together with the realized frozen-class size, mean paired pulls, and final unresolved gap. No minimum-budget-only outcome occurred. (b) Certified budgets are summarized among the 80 dual certificates. (c) Adaptive and uniform allocation use the same confidence sequences, $(\eta,\delta)$, and stopping predicates at a matched 91.4\% at-least-optimality rate. (d) Resolution is stratified by the completed-audit empirical top-two arm gap.}
\label{fig:direct-signed-cmr-audit}
\end{figure*}

Across all panels, 117/128 instances (91.4\%) receive at least an optimality certificate, while 80/128 (62.5\%) also resolve the separate minimum-budget certificate. Among dual certificates, 73.8\% certify a one-factor budget, 22.5\% a two-factor budget, and 3.8\% a three-factor budget.

At the same 91.4\% at-least-optimality certificate rate, adaptive signed-CMR allocation reduces median paired pulls from 518 to 359, a 30.7\% rollout reduction. The dual-certificate rate falls from 83\% for empirical top-arm gaps of at least 0.15 to 27\% below 0.05, concentrating unresolved instances among small-gap cases at the registered cap.
}

\begin{figure}[!htbp]
\centering
\includegraphics[draft=false,width=\linewidth]{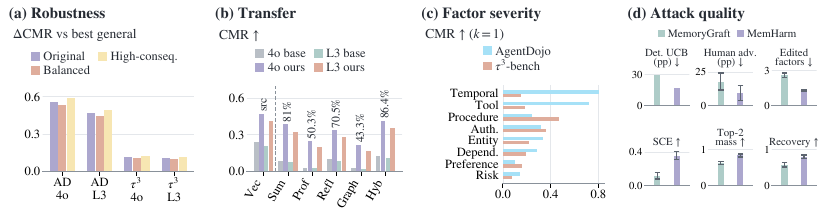}
\caption{\textbf{Severity structure and generalization.} Panels (a--d) summarize objective robustness, cross-memory transfer, semantic-factor severity, and attack-quality diagnostics. Whiskers denote 95\% CIs where available; transfer labels report macro retention.}
\label{fig:severity-structure}
\end{figure}

\subsection{Mechanism: Consequence and Memory Realization}
\label{subsec:mechanism-validation}
The preregistered panel examines the consequence--realization account in Section~\ref{sec:theory}. The directional bound is nonvacuous in 19/30 positive-consequence standard-memory cells, whereas strict-schema controls have $V_{g,K}=0$ (Appendix~\ref{app:mechanism-diagnostics}, Table~\ref{tab:mechanism-validation}).

Removing the selected factor changes CMR by $-0.34$, compared with $-0.02$ for a matched irrelevant factor; the respective 95\% intervals exclude and include zero (Appendix~\ref{app:mechanism-diagnostics}, Table~\ref{tab:mechanism-validation}). This intervention supports a factor-specific contribution to the observed harm; single-factor and pipeline breakdowns appear in Appendix~\ref{app:factor-breakdown}, Figure~\ref{fig:factor-analysis}.

\newcommand{\ccmaExactMechanismTable}{%
\begin{center}
\begin{minipage}{\linewidth}
\captionsetup{hypcap=false}
\captionof{table}{\textbf{Mechanism validation.} (a) A fresh preregistered directional panel contains 42 completed aggregated cells; the secondary utility association excludes the six strict-schema cells ($n=36$). (b) Factor-removal changes are after minus before with task-cluster bootstrap 95\% intervals.}
\label{tab:mechanism-validation}
\centering
\footnotesize
\textbf{(a) Directional accounting and secondary diagnostic}\par
\vspace{0.3em}
\resizebox{0.94\linewidth}{!}{%
\begin{tabular}{lccc}
\toprule
\textbf{Test} & \textbf{Estimate} & \textbf{Uncertainty / denominator} & \textbf{$p$} \\
\midrule
Directional lower bound satisfied & 42/42 cells & 42 complete cells & -- \\
Nonvacuous bound, positive-consequence cells & 63.3\% & 19/30 cells & -- \\
OLS slope on $\mathrm{CU}_g$ conditional on factor / $\Psi$ bin & 0.64 & approx.\ 95\% CI $[0.20,1.08]$; $n=36$ & $<0.01$ \\
Strict-schema control $V_{g,K}=0$ & 6/6 cells & 6 strict-schema cells & -- \\
\bottomrule
\end{tabular}%
}
\par\smallskip
\parbox{0.94\linewidth}{\scriptsize The association interval is approximate. The 63.3\% nonvacuity rate exceeds the preregistered 50\% criterion.}

\vspace{0.7em}
\textbf{(b) Causal-path ablation}\par
\vspace{0.3em}
\resizebox{0.70\linewidth}{!}{%
\begin{tabular}{lcc}
\toprule
\textbf{Intervention} & \textbf{Change in CMR} & \textbf{95\% CI conclusion} \\
\midrule
Remove selected semantic factor & $-0.34$ & Entirely below zero \\
Remove matched irrelevant factor & $-0.02$ & Includes zero \\
\bottomrule
\end{tabular}%
}
\end{minipage}
\end{center}
}

With vector-memory selection frozen before any target rollout, attacks retain substantial CMR in summary, reflection, and hybrid memories, but transfer weakens in schema-constrained profile and graph stores (Figure~\ref{fig:severity-structure}(b); Appendix~\ref{app:defense-transfer}, Figure~\ref{fig:transfer}). Target outcomes enter neither candidate generation nor selection. This pattern is consistent with transferable semantic corruption whose severity depends on what the target memory can realize.

\begin{figure}[htbp]
\centering
\includegraphics[draft=false,width=\linewidth]{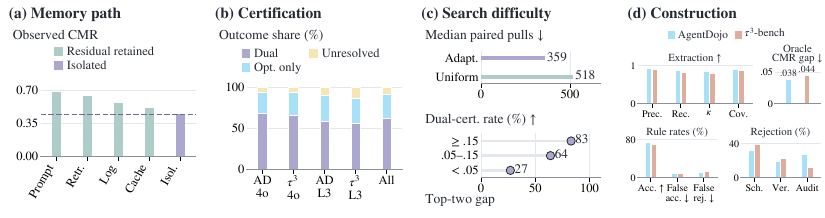}
\caption{\textbf{Identification and reliability.} Panels (a--d) summarize memory-path identification, class-conditional certification, certificate difficulty, and construction validity. Whiskers denote 95\% CIs where available.}
\label{fig:severity-reliability}
\end{figure}

\subsection{Validating the Attack-Severity Claim}
\label{subsec:severity-validity}
Residual-channel controls support memory-only isolation (Appendix~\ref{app:sensitivity}, Figure~\ref{fig:memory-only-sensitivity}); 91.4\% of attacks receive optimality certificates, with adaptive allocation and gap effects (Figure~\ref{fig:severity-reliability}(a--c); Appendix~\ref{app:minimality-cert}, Figure~\ref{fig:direct-signed-cmr-audit}). Construction audits report extractor agreement (Cohen's $\kappa$), oracle gap, and rejection rates (Figure~\ref{fig:severity-reliability}(d); Appendix~\ref{app:ccma-construction-validation}, Figure~\ref{fig:factor-coverage} and Table~\ref{tab:detectors}). Gemini and reweighting results appear in Appendices~\ref{app:backbone-ablation}--\ref{app:severity-reweighting}, Tables~\ref{tab:gemini-35-flash-agentdojo}--\ref{tab:severity-reweighting}; defense and blinded-retention audits appear in Appendices~\ref{app:defense-transfer}--\ref{app:retention-recovery} (Figures~\ref{fig:defense} and~\ref{fig:stealth}). Together, these results support memory mediation and class-conditional selection reliability.

\newcommand{\ccmaDefenseTable}{%
\begin{table*}[htbp]
\caption{Defense-stress results for GPT-4o, reported as mean CMR (lower is better for the defender) $\pm$ standard error across five independent seeds and equal-weight macro-averaged over AgentDojo and $\tau^3$-bench. No-defense entries are method-specific baselines; the common-support attack-effectiveness comparison appears in Table~\ref{tab:main-effectiveness}, with exact values in Table~\ref{tab:main-results}. Benign drop is the clean-memory utility loss induced by each defense.}
\label{tab:defense}
\centering
\setlength{\tabcolsep}{3.0pt}
\scriptsize
\resizebox{\textwidth}{!}{%
\begin{tabular}{lcccccccc}
\toprule
\textbf{Defense}
& \textbf{AgentPoison}
& \textbf{MINJA}
& \textbf{InjecMEM}
& \textbf{MemoryGraft}
& \textbf{MemPoison}
& \textbf{DrunkAgent}
& \textbf{MemHarm}
& \textbf{Benign drop} \\
\midrule

No defense
& $0.068 \pm 0.031$
& $0.044 \pm 0.028$
& $0.061 \pm 0.034$
& $0.185 \pm 0.030$
& $0.116 \pm 0.033$
& $0.060 \pm 0.031$
& $\mathbf{0.479 \pm 0.035}$
& $0.000 \pm 0.000$ \\

Input injection filter
& $0.048 \pm 0.027$
& $0.041 \pm 0.029$
& $0.040 \pm 0.026$
& $0.184 \pm 0.031$
& $0.112 \pm 0.032$
& $0.049 \pm 0.029$
& $\mathbf{0.442 \pm 0.034}$
& $0.016 \pm 0.011$ \\

Contradiction detection
& $0.037 \pm 0.024$
& $0.029 \pm 0.027$
& $0.031 \pm 0.026$
& $0.145 \pm 0.023$
& $0.079 \pm 0.025$
& $0.034 \pm 0.025$
& $\mathbf{0.377 \pm 0.032}$
& $0.029 \pm 0.021$ \\

Provenance verification
& $0.018 \pm 0.019$
& $0.017 \pm 0.023$
& $0.012 \pm 0.017$
& $0.102 \pm 0.021$
& $0.048 \pm 0.018$
& $0.017 \pm 0.020$
& $\mathbf{0.303 \pm 0.029}$
& $0.087 \pm 0.052$ \\

\shortstack[l]{Human confirmation for\\high-utility factors}
& $0.016 \pm 0.018$
& $0.015 \pm 0.022$
& $0.010 \pm 0.016$
& $0.096 \pm 0.020$
& $0.044 \pm 0.019$
& $0.015 \pm 0.019$
& $\mathbf{0.323 \pm 0.031}$
& $0.042 \pm 0.026$ \\

MemAudit
& $0.011 \pm 0.015$
& $0.011 \pm 0.018$
& $0.007 \pm 0.014$
& $0.082 \pm 0.019$
& $0.036 \pm 0.016$
& $0.011 \pm 0.016$
& $\mathbf{0.261 \pm 0.027}$
& $0.036 \pm 0.022$ \\
\midrule

\shortstack[l]{Combined provenance + confirmation\\+ causal audit}
& $0.008 \pm 0.013$
& $0.009 \pm 0.016$
& $0.005 \pm 0.012$
& $0.067 \pm 0.017$
& $0.026 \pm 0.014$
& $0.009 \pm 0.014$
& $\mathbf{0.210 \pm 0.021}$
& $0.059 \pm 0.033$ \\

\bottomrule
\end{tabular}%
}
\end{table*}
\FloatBarrier

Table~\ref{tab:defense} reports the observed defense-stress outcomes. For MemHarm, the input-injection filter reduces mean CMR from $0.479$ to $0.442$, whereas MemAudit reduces it to $0.261$ and the combined provenance, confirmation, and causal-audit defense reduces it to $0.210$. The combined defense attains the lowest mean CMR for every attack method, with a mean benign drop of $0.059\pm0.033$.
}

\newcommand{\ccmaDefenseFigure}{%
\begin{figure*}[t]
\centering
\includegraphics[draft=false,width=\textwidth]{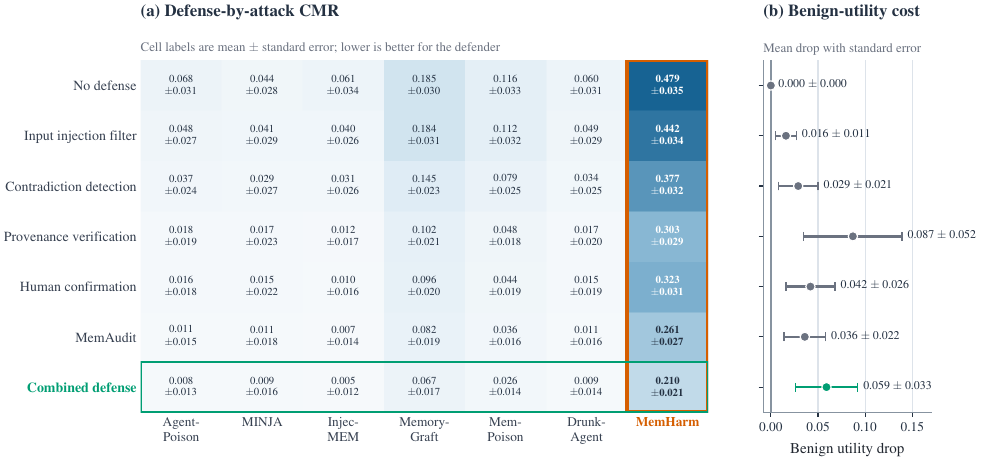}
\caption{Defense-stress results for GPT-4o, equal-weight macro-averaged over AgentDojo and $\tau^3$-bench across five independent seeds. (a) Each cell reports mean CMR $\pm$ standard error; lower is better for the defender. The orange outline marks MemHarm and the green outline marks the combined provenance, confirmation, and causal-audit defense. (b) The aligned benign-utility cost is reported as mean drop $\pm$ standard error.}
\label{fig:defense}
\end{figure*}

Appendix~\ref{app:defense-transfer} reports the observed defense-stress outcomes. For MemHarm, the input-injection filter reduces mean CMR from $0.479$ to $0.442$, whereas MemAudit reduces it to $0.261$ and the combined defense reduces it to $0.210$. The combined defense attains the lowest mean CMR for every attack method, with a mean benign drop of $0.059\pm0.033$.
}

\newcommand{\ccmaTransferTable}{%
\paragraph{Cross-memory transfer protocol.}
For each benchmark--backbone block, we evaluate 48 held-out source tasks with five independent seeds. MemHarm is selected using source-memory data only and frozen before any target-memory rollout; target outcomes never enter candidate generation or selection. Confidence intervals use a task-cluster bootstrap that resamples all five seeds of a task together. Taking vector memory as the registered source, we report $R_{\mathrm{transfer}}=\CMR_{\mathrm{target}}/\CMR_{\mathrm{source}}$.

\begin{table}[htbp]
\caption{Cross-memory transfer macro-averaged over two benchmarks and two backbones. Each block contains 48 held-out source tasks and five independent seeds. CMR brackets are task-cluster bootstrap 95\% confidence intervals; retention ratios use the corresponding point estimates.}
\label{tab:transfer}
\centering
\resizebox{\linewidth}{!}{%
\begin{tabular}{lcccc}
\toprule
\textbf{Target memory} & \textbf{Vector-only baseline CMR} & $\boldsymbol{R_{\mathrm{transfer}}}$ & \textbf{Frozen source-memory MemHarm CMR} & $\boldsymbol{R_{\mathrm{transfer}}}$ \\
\midrule
Vector & $0.224\ [0.179, 0.269]$ & 1.000 & $\mathbf{0.441\ [0.389, 0.493]}$ & \textbf{1.000} \\
Summary & $0.083\ [0.050, 0.116]$ & 0.371 & $\mathbf{0.357\ [0.309, 0.405]}$ & \textbf{0.810} \\
Profile & $0.031\ [0.014, 0.048]$ & 0.138 & $\mathbf{0.222\ [0.183, 0.261]}$ & \textbf{0.503} \\
Reflection & $0.095\ [0.060, 0.130]$ & 0.424 & $\mathbf{0.311\ [0.267, 0.355]}$ & \textbf{0.705} \\
Graph & $0.022\ [0.008, 0.036]$ & 0.098 & $\mathbf{0.191\ [0.156, 0.226]}$ & \textbf{0.433} \\
Hybrid & $0.119\ [0.079, 0.159]$ & 0.531 & $\mathbf{0.381\ [0.333, 0.429]}$ & \textbf{0.864} \\
\bottomrule
\end{tabular}%
}
\end{table}

Table~\ref{tab:transfer} measures transfer after source-only selection is frozen. MemHarm retains 81.0\% of source CMR in summary memory, 70.5\% in reflection memory, and 86.4\% in hybrid memory; profile and graph transfer are lower, consistent with stricter schemas and typed validation.
}

\newcommand{\ccmaTransferFigure}{%
\paragraph{Cross-memory transfer protocol.}
For each benchmark--backbone block, we evaluate 48 held-out source tasks with five independent seeds. MemHarm is selected using source-memory data only and frozen before any target-memory rollout; target outcomes never enter candidate generation or selection. Confidence intervals use a task-cluster bootstrap that resamples all five seeds of a task together. Taking vector memory as the registered source, we report $R_{\mathrm{transfer}}=\CMR_{\mathrm{target}}/\CMR_{\mathrm{source}}$.

\begin{figure*}[t]
\centering
\includegraphics[draft=false,width=\textwidth]{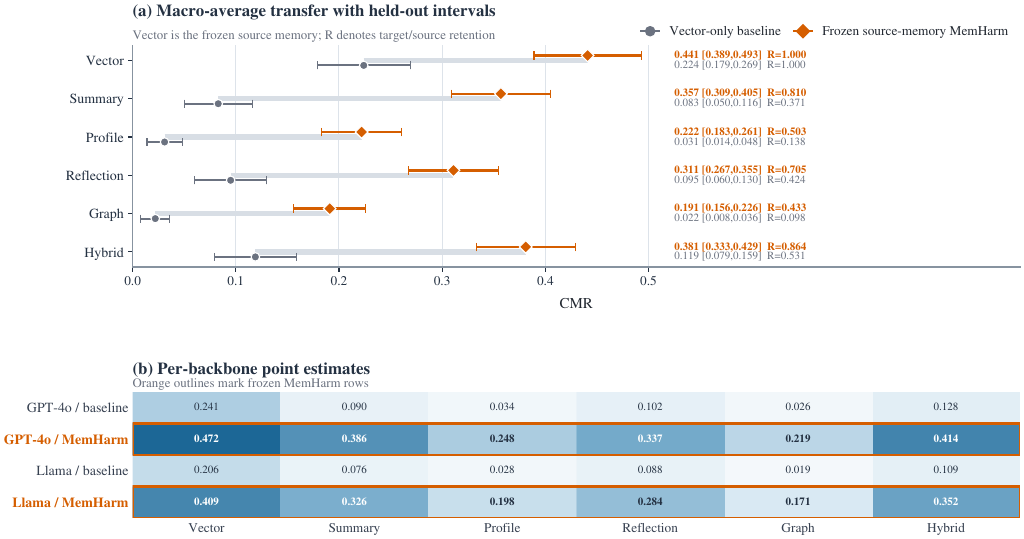}
\caption{Cross-memory transfer after source-only selection is frozen. (a) Macro-averages over two benchmarks and two backbones; each block contains 48 held-out source tasks and five independent seeds. Whiskers are task-cluster bootstrap 95\% confidence intervals and direct labels include the retention ratio $R_{\mathrm{transfer}}$. (b) Per-backbone CMR point estimates underlying the macro-average; orange outlines mark source-memory MemHarm rows.}
\label{fig:transfer}
\end{figure*}

Appendix~\ref{app:defense-transfer} shows that MemHarm retains 81.0\% of source CMR in summary memory, 70.5\% in reflection memory, and 86.4\% in hybrid memory; profile and graph transfer are lower, consistent with stricter schemas and typed validation.
}

\newcommand{\ccmaStealthTable}{%
\paragraph{Held-out detectability, retention, and recovery audits.}
For each method, automated detectors use at least 240 attacked writes and 240 matched-clean writes disjoint from construction, optimization, and detector calibration. Every human-rated item receives at least three independent blinded judgments. Human flagging asks whether an item is attack-derived; BMAR asks whether a single item should be retained under an ordinary maintenance decision. Recovery is judged by a separate cohort that does not see the method name or CMR and records whether the recovered explanation identifies the registered causal factor path.

\begin{table*}[htbp]
\caption{Held-out stealth, sparsity, recovery, and retention audits. (a) MemoryGraft and MemHarm are compared across both benchmarks and primary backbones on each method's eligible support. $\Delta_{\Hcal}$ is the simultaneous UCB over the frozen detector family; bracketed values are task-cluster bootstrap 95\% confidence intervals. Recovery uses a separate blinded annotator cohort. (b) The retention audit uses 240 MemHarm writes and 240 matched-clean writes, at least three reviewers per item, task-level bootstrap intervals, and the task as the independent unit.}
\label{tab:stealth}
\centering
\textbf{(a) Stealth, sparsity, and causal faithfulness.}\par
\vspace{0.2em}
\resizebox{\textwidth}{!}{%
\begin{tabular}{lcccccc}
\toprule
\textbf{Method} & $\boldsymbol{\Delta_{\Hcal}}$ \textbf{UCB $\downarrow$} & \textbf{Human flagging adv. $\downarrow$} & \textbf{Edit sparsity $\downarrow$} & \textbf{SCE $\uparrow$} & \textbf{Top-2 mass $\uparrow$} & \textbf{Recovery $\uparrow$} \\
\midrule
MemoryGraft & 0.296 & $0.181\ [0.118, 0.244]$ & $2.62\ [2.42, 2.82]$ & $0.119\ [0.081, 0.157]$ & $0.63\ [0.58, 0.68]$ & $0.58\ [0.52, 0.64]$ \\
\textbf{MemHarm (ours)} & \textbf{0.176} & $\mathbf{0.096\ [0.040, 0.151]}$ & $\mathbf{1.34\ [1.27, 1.41]}$ & $\mathbf{0.357\ [0.309, 0.404]}$ & $\mathbf{0.84\ [0.80, 0.88]}$ & $\mathbf{0.80\ [0.75, 0.85]}$ \\
\bottomrule
\end{tabular}%
}

\vspace{0.7em}
\textbf{(b) Contextual apparent benignness.}\par
\label{tab:bmar-results}
\vspace{0.2em}
\begin{tabular}{lc}
\toprule
\textbf{Metric} & \textbf{Estimate (95\% CI)} \\
\midrule
MemHarm $\BMAR_{\geq4}$ & $0.804\ [0.752, 0.851]$ \\
Matched-clean $\BMAR_{\geq4}$ & $0.887\ [0.846, 0.921]$ \\
Mean MPS attack--clean appearance gap & $-0.11\ [-0.17, -0.05]$ \\
Krippendorff's $\alpha$ & $0.73\ [0.68, 0.78]$ \\
\bottomrule
\end{tabular}
\end{table*}

MemHarm has lower detector and human-flagging advantages than MemoryGraft, edits fewer factors, and achieves higher stealth-discounted regret, top-two causal mass, and recovery. Reviewers retain 80.4\% of MemHarm writes and 88.7\% of matched-clean writes; the mean appearance gap is $-0.11$ with Krippendorff's $\alpha=0.73$.
}

\newcommand{\ccmaStealthFigure}{%
For each method, automated detectors use at least 240 attacked writes and 240 matched-clean writes disjoint from construction, optimization, and detector calibration. Every human-rated item receives at least three independent blinded judgments. Human flagging asks whether an item is attack-derived; BMAR asks whether a single item should be retained under an ordinary maintenance decision. Recovery is judged by a separate cohort that does not see the method name or CMR and records whether the recovered explanation identifies the registered causal factor path.

MemHarm has lower detector and human-flagging advantages than MemoryGraft, edits fewer factors, and achieves higher stealth-discounted regret, top-two causal mass, and recovery. Reviewers retain 80.4\% of MemHarm writes and 88.7\% of matched-clean writes; the mean appearance gap is $-0.11$ with Krippendorff's $\alpha=0.73$.

\begin{figure*}[t]
\centering
\includegraphics[draft=false,width=0.92\textwidth]{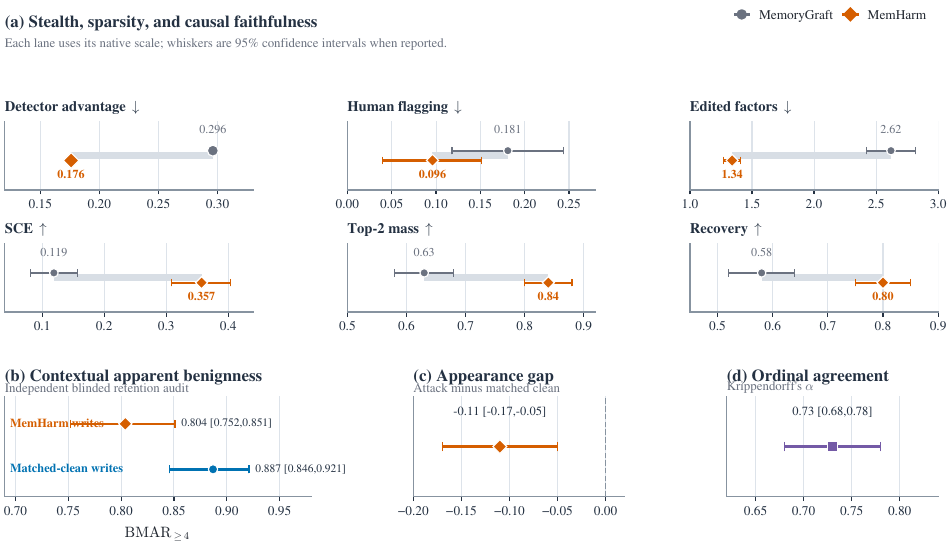}
\caption{Held-out stealth, sparsity, recovery, and retention audits. (a) MemoryGraft and MemHarm are compared on six metrics using each metric's native scale; whiskers denote task-cluster bootstrap 95\% confidence intervals and $\Delta_{\Hcal}$ is a simultaneous UCB. (b) The retention audit uses 240 MemHarm writes and 240 matched-clean writes, at least three reviewers per item, and task-level bootstrap intervals. (c) The attack--clean appearance gap. (d) Bootstrap interval for ordinal agreement measured by Krippendorff's $\alpha$.}
\label{fig:stealth}
\label{fig:bmar-results}
\end{figure*}
}

\newcommand{\ccmaFactorTable}{%
\begin{table}[htbp]
\caption{Factor and pipeline breakdown for MemHarm on the two primary benchmarks, averaged over GPT-4o and Llama-3.1-70B-Instruct. The top block reports point-estimate CMR for single-factor edits; the bottom block reports rejection and audit-failure rates.}
\label{tab:factor-analysis}
\begin{center}
\resizebox{\linewidth}{!}{
\begin{tabular}{lcc}
\toprule
\textbf{Semantic factor} & \textbf{AgentDojo} & \textbf{$\tau^3$-bench} \\
\midrule
\multicolumn{3}{l}{\emph{Per-factor CMR contribution (MemHarm restricted to single-factor edits, $k{=}1$)}} \\
\midrule
Entity binding & 0.336 & 0.218 \\
Preference shift & 0.097 & 0.163 \\
Procedure prior & 0.244 & \textbf{0.472} \\
Temporal priority & \textbf{0.806} & 0.151 \\
Tool affordance & \textbf{0.723} & 0.184 \\
Authorization & 0.318 & \textbf{0.361} \\
Dependency & 0.286 & 0.195 \\
Risk threshold & 0.142 & 0.079 \\
\midrule
\multicolumn{3}{l}{\emph{Pipeline failure rates}} \\
\midrule
Strict-schema profile rejection rate & 31.7\% & 38.9\% \\
Verifier-rejected candidate rate & 18.6\% & 22.4\% \\
Human-audit rejection rate & 26.4\% & 11.4\% \\
\bottomrule
\end{tabular}}
\end{center}
\end{table}

Table~\ref{tab:factor-analysis} summarizes where single-factor CMR and pipeline rejection concentrate. The complete directional and intervention results appear in Appendix~\ref{app:mechanism-diagnostics}.
}

\newcommand{\ccmaFactorFigure}{%
\begin{center}
\captionsetup{hypcap=false}
\centering
\includegraphics[draft=false,width=0.90\textwidth]{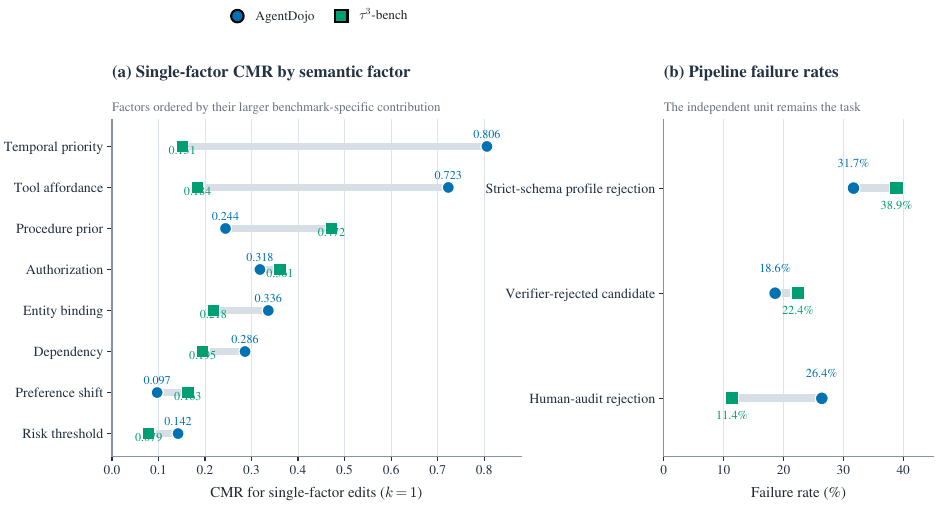}
\captionof{figure}{Factor and pipeline breakdown, averaged over both backbones. (a) Single-factor CMR by benchmark, ordered by the larger point estimate. (b) Pipeline rejection and audit-failure rates.}
\label{fig:factor-analysis}
\end{center}

Appendix~\ref{app:factor-breakdown} summarizes where single-factor CMR and pipeline rejection concentrate.
}

\FloatBarrier

\section{Conclusion and Limitations}
\label{sec:scope-conclusion}

We introduced MemHarm to shift persistent-memory attack design from binary success toward downstream severity. By formalizing severity as counterfactual memory regret (CMR), MemHarm searches for persistent states that cause greater downstream harm relative to clean memory, while providing class-conditional guarantees over a frozen attack space. Empirically, severity-guided selection consistently identifies more harmful persistent states than success-oriented or matched baselines while remaining strong under conventional attack success.

The resulting claims are deliberately scoped to this formulation. Certification applies to the declared finite edit class (\S\ref{subsec:fixed-confidence}; Appendix~\ref{app:minimality-cert}), and attack construction uses resettable isolated copies with paired-loss feedback, while deployment requires only a normal-interface memory write (\S\ref{subsec:memory-mediated-severity}). CMR is likewise defined with respect to a declared downstream task distribution and operational loss, so its value should be interpreted within that evaluation setting. Broader settings and extensions are discussed in Appendices~\ref{app:limitations}--\ref{app:future-work}.

\clearpage
\subsection*{AI Use Statement}
\label{sec:ai-use-statement}
Generative AI was used for language polishing, including grammar, clarity, and stylistic editing; to assist with code development and debugging; and to support brainstorming of figure concepts and visual design ideas. LLMs were also used as experimental backbones and as components of the experimental pipeline as described in Sections~\ref{sec:ccma-method} and~\ref{sec:experiments} and Appendix~\ref{app:implementation}. All AI-assisted text, code, and analyses were critically reviewed and verified against the underlying sources, data, and implementations. The authors reviewed all AI-assisted content and take full responsibility for the final manuscript.

\subsection*{Ethics Statement}
\label{sec:ethics-statement}
MemHarm is a dual-use memory attack against persistent agent state. Experiments use isolated benchmark environments and locally deployed OpenClaw and Hermes instances, never production services or live accounts. We release aggregate results, validators, and audit protocols but not live-service exploit code or deployment instructions. Human annotators completed the blinded appearance and causal-recovery judgments in Appendices~\ref{app:contextual-benignness} and~\ref{app:retention-recovery}. They were recruited through Prolific and required to be at least 18 years old, fluent in English, and have a platform approval rate of at least 95\%. Compensation was calibrated to expected completion time at an effective rate of at least US\$15 per hour, and participants provided electronic informed consent before beginning. We collected no names, email addresses, or other additional personally identifiable information; retained records contain only pseudonymous worker IDs and judgments. The released validators and audit protocols are intended for controlled defensive audits of memory-write provenance, validation, and confirmation.

\subsection*{Reproducibility Statement}
\label{sec:reproducibility-statement}
Sections~\ref{sec:audit-cmr} and~\ref{sec:experiments} specify the estimand, candidate freezing, paired evaluation, support matching, and primary metrics. The appendix reports the confidence sequences, memory contracts, task splits, validators, statistical tests, sample sizes, complete results, and audit protocols; Appendix~\ref{app:reporting-template}, Figure~\ref{fig:reporting-template} specifies the per-attack record. We release task-level statistics, validators, and the registered protocol for independent verification and protocol-level reproduction; operational exploit code is excluded as described in the Ethics Statement above.

\clearpage
\bibliography{iclr2027_conference}
\bibliographystyle{iclr2027_conference}

\clearpage
\appendix
\input{appendix_contents.tex}

\section{Extended Comparison with Prior Memory Attacks}
\label{app:prior-work-comparison}

Table~\ref{tab:prior-attacks-full} compares the attack variable, memory path, and evaluation target of MemHarm with representative prior attacks.

\begin{table*}[!t]
\caption{\textbf{Qualitative comparison with prior memory attacks} using each source paper's terminology and evaluation protocol.}
\label{tab:prior-attacks-full}
\centering
\footnotesize
\setlength{\tabcolsep}{2.5pt}
\renewcommand{\arraystretch}{1.08}
\begin{tabular}{>{\raggedright\arraybackslash}p{0.13\textwidth}>{\raggedright\arraybackslash}p{0.12\textwidth}>{\raggedright\arraybackslash}p{0.19\textwidth}>{\raggedright\arraybackslash}p{0.15\textwidth}>{\raggedright\arraybackslash}p{0.16\textwidth}>{\raggedright\arraybackslash}p{0.15\textwidth}}
\toprule
\textbf{Method} & \textbf{Attack variable} & \textbf{Reported insertion / retention path} & \textbf{Role of retrieval in the reported mechanism} & \textbf{Reported detectability / plausibility framing} & \textbf{Evaluation target} \\
\midrule
PoisonedRAG & Malicious passages & Retrieval-corpus or index content; agent-memory state is not the primary target & Central to the attack & Perplexity-based and duplicate-filter defenses evaluated & Attacker-chosen answer for an attacker-chosen target question \\
AgentPoison & Trigger-conditioned poisoned demonstrations & Poisoned long-term-memory or knowledge-base entries & Central in the reported retrieval-equipped configurations & Trigger coherence; benign-task behavior also measured & Triggered target action and end-to-end adversarial impact \\
MINJA & Query-induced malicious records with bridging steps & Records induced through query-only interaction and retained by the agent & Central to later activation; query-only describes the insertion route & Query-only interaction constraint & Targeted malicious reasoning on later victim queries \\
InjecMEM & Topic anchor paired with an instruction payload & A record induced in one interaction and retained through the memory path & Retriever-agnostic anchor designed for topic-conditioned later retrieval & Topic-conditioned retrieval & Pre-specified output for related queries \\
MemoryGraft & Poisoned successful experiences or procedure templates & Ingestion artifacts that induce persistent poisoned experience records & Experience retrieval is central & Successful-looking traces & Persistent behavioral drift on semantically similar tasks \\
MemPoison & Dialogue-delivered trigger--payload relation & A triggerable backdoor retained through selective memory processing & Depends on the memory extraction/retrieval pipeline & Selective-memory survival & Trigger-conditioned later responses \\
DrunkAgent & Adversarial target-item description and memory-update strategy & A modified description perturbs the target item agent's dynamic memory and impedes later updates & Architecture-dependent; retrieval is explicit in AgentRAG but not universal across AgentCF, AgentRAG, and AgentSEQ & Low-perplexity descriptions; limited change in overall recommendation performance & Target-item promotion, measured by HR@$K$ and NDCG@$K$ \\
\textbf{MemHarm} & Frozen typed semantic edit and concrete write & Normal memory-write interface in an isolated evaluation copy & Uses each memory contract's native read interface across six backends & Held-out detector-family tests and causal isolation & Paired downstream task regret \\
\bottomrule
\end{tabular}
\par\vspace{0.6em}
\begin{minipage}{\linewidth}
\normalsize
\textbf{Construction requirements.}
MemHarm's offline search requires resettable isolated copies and paired-loss feedback; deployment uses one normal-interface write. Table~\ref{tab:construction-overhead} reports measured paired-rollout, token, and wall-clock costs against no-feedback and uniform-search controls on the same instances. Cross-paper costs are not directly ranked because models, environments, and accounting conventions differ.
\end{minipage}
\end{table*}

\section{Paired Intervention and Sampling Assumptions}
\label{app:paired}

The paired CMR estimator uses four design conditions. \emph{Consistency} requires that a copy assigned write $x$ realizes state $U_g(S_0,x)$ and the corresponding downstream loss. \emph{No cross-copy interference} requires that one arm's rollouts do not alter another arm's memory, tool database, vector index, cache, or task state. The \emph{memory-only intervention} removes the original artifact from prompt, inbox view, non-target retrieval stores, tool log, and auxiliary cache after writing, unless that content is part of the persistent memory object being tested. \emph{Conditional mean stability} requires that, after conditioning on the frozen class and past adaptive allocation, each fresh paired rollout for arm $a$ has bounded difference in $[-L_{\max},L_{\max}]$ and conditional mean $J_g(a;X)$. Independent rollout seeds satisfy this condition; common random numbers may be used to reduce variance.

For a nonadaptively selected arm $a=(e,x)$, the paired estimator
\begin{equation}
 \widehat J_g(a;X)=\frac{1}{n}\sum_{i=1}^n
 \left[L_g(S_g^x;Q_i,\omega_i^{\mathrm{edit}})-L_g(S_g^0;Q_i,\omega_i^{\mathrm{clean}})\right]
\end{equation}
is unbiased for Equation~\ref{eq:cmr} under any valid coupling of $(\omega_i^{\mathrm{edit}},\omega_i^{\mathrm{clean}})$ with the correct branch marginals. Adaptive selection is certified using the simultaneous intervals in Appendix~\ref{app:minimality-cert}, rather than a post-selection interval for the empirical winner.

\section{Extended Directional Analysis}
\label{app:directional-analysis}

Fix a true task-evaluable factor value $C=c$ under context $H$. For factors with an executable evaluator reference, let $\lambda_Q(v\mid c,H)\in[0,L_{\max}]$ be the loss of a registered reference behavior that treats the factor as $v$ while task truth remains $c$. The reference map is frozen before attack search from a benchmark rule, database transition, tool constraint, or unit-test oracle. Define
\begin{equation}
\label{eq:directional-consequence}
 \kappa(a\mid c,H)
 =\E_Q\!\left[\lambda_Q(v_a\mid c,H)-\lambda_Q(c\mid c,H)\right].
\end{equation}
For multi-factor arms, $c$ and $v_a$ denote the corresponding vectors.

For arm $a=(e,x)$, the clean and edited realization biases are
\begin{align}
\label{eq:branch-bias-clean}
 \rho_g^{\mathrm{clean}}(c,H)
 &=\E_{Q,\omega^{\mathrm{clean}}}\!\left[L_g(S_g^0;Q,\omega^{\mathrm{clean}})-\lambda_Q(c\mid c,H)\right],\\
\label{eq:branch-bias-attack}
 \rho_g^{\mathrm{edit}}(a;c,H)
 &=\E_{Q,\omega^{\mathrm{edit}}}\!\left[L_g(S_g^x;Q,\omega^{\mathrm{edit}})-\lambda_Q(v_a\mid c,H)\right],
\end{align}
with absolute realization errors
\begin{align}
\label{eq:clean-reference-error}
 \varepsilon_g^{\mathrm{clean}}(c,H)
 &=\E_{Q,\omega^{\mathrm{clean}}}\!\left|L_g(S_g^0;Q,\omega^{\mathrm{clean}})-\lambda_Q(c\mid c,H)\right|,\\
\label{eq:attack-reference-error}
 \varepsilon_g^{\mathrm{edit}}(a;c,H)
 &=\E_{Q,\omega^{\mathrm{edit}}}\!\left|L_g(S_g^x;Q,\omega^{\mathrm{edit}})-\lambda_Q(v_a\mid c,H)\right|.
\end{align}
Equation~\ref{eq:exact-directional-decomposition} therefore implies
\begin{equation}
\label{eq:per-arm-directional-bound}
 J_g(a;X)\geq
 \kappa(a\mid c,H)-\varepsilon_g^{\mathrm{edit}}(a;c,H)-\varepsilon_g^{\mathrm{clean}}(c,H).
\end{equation}
For a class admitting a non-clean arm, define the mechanism-conditioned realizable attack potential
\begin{equation}
\label{eq:realizable-potential}
 \Psi_{g,K}(c,H)
 =\max_{a\in\Acal_K(X)\setminus\{a_0\}}
 \left\{\kappa(a\mid c,H)-\varepsilon_g^{\mathrm{edit}}(a;c,H)\right\}.
\end{equation}
Since the zero-edit arm satisfies $J_g(a_0;X)=0$,
\begin{equation}
\label{eq:duality-bound}
 V_{g,K}(X)
 \geq\max\left\{0,\Psi_{g,K}(c,H)-\varepsilon_g^{\mathrm{clean}}(c,H)\right\}.
\end{equation}

\begin{corollary}[Conditional utility--attackability bound]
\label{cor:utility-amplification}
Let a factor-blind or no-memory baseline have clean reference error $\varepsilon_\emptyset^{\mathrm{clean}}(c,H)$, and define
\begin{equation}
\label{eq:utility-gain}
 \Util_g(c,H)=\varepsilon_\emptyset^{\mathrm{clean}}(c,H)-\varepsilon_g^{\mathrm{clean}}(c,H).
\end{equation}
Then
\begin{equation}
\label{eq:utility-risk-corollary}
 V_{g,K}(X)
 \geq\max\left\{0,\Psi_{g,K}(c,H)-\varepsilon_\emptyset^{\mathrm{clean}}(c,H)+\Util_g(c,H)\right\}.
\end{equation}
Thus utility tightens the bound within a fixed realizable-potential regime, while realizable potential determines whether that utility supports high CMR.
\end{corollary}

\section{Proofs}
\label{app:proofs}

\subsection{CMR Decomposition and Directional Bound}
For arm $a=(e,x)$, add and subtract the two reference losses:
\begin{align}
J_g(a;X)
&=\E[L_g(S_g^x)-L_g(S_g^0)]\\
&=\E[\lambda_Q(v_a\mid c,H)-\lambda_Q(c\mid c,H)]\\
&\quad+\E[L_g(S_g^x)-\lambda_Q(v_a\mid c,H)]\\
&\quad-\E[L_g(S_g^0)-\lambda_Q(c\mid c,H)]\\
&=\kappa(a\mid c,H)+\rho_g^{\mathrm{edit}}(a;c,H)-\rho_g^{\mathrm{clean}}(c,H),
\end{align}
which proves Equation~\ref{eq:exact-directional-decomposition}. Since $\rho_g^{\mathrm{edit}}(a;c,H)\geq-\varepsilon_g^{\mathrm{edit}}(a;c,H)$ and $-\rho_g^{\mathrm{clean}}(c,H)\geq-\varepsilon_g^{\mathrm{clean}}(c,H)$, Equation~\ref{eq:per-arm-directional-bound} follows. Maximizing over non-clean arms and using $J_g(a_0;X)=0$ gives Equation~\ref{eq:duality-bound}. \hfill$\square$

\paragraph{Proof of Corollary~\ref{cor:utility-amplification}.}
Substitute $\varepsilon_g^{\mathrm{clean}}=\varepsilon_\emptyset^{\mathrm{clean}}-\Util_g$ into Equation~\ref{eq:duality-bound}. \hfill$\square$

\subsection{Proof of Proposition~\ref{prop:fixed-confidence}}
\label{app:proof-selection}
On the simultaneous confidence event, $J_g(a;X)\in[\LCB_t(a),\UCB_t(a)]$ for all arms and all times. At stopping,
\begin{equation}
 V_{g,K}(X)
 \leq\max_a\UCB_t(a)
 \leq\LCB_t(\widehat a_t)+\eta
 \leq J_g(\widehat a_t;X)+\eta.
\end{equation}
\hfill$\square$

\subsection{Proof of Proposition~\ref{prop:causal-locus-nonidentifiability}}
\label{app:proof-attribution}
Let $F\sim\mathrm{Unif}\{i,j\}$ and $O\mid(F=f)\sim P_f^O$. Equal-prior attribution is binary hypothesis testing between $P_i^O$ and $P_j^O$, for which
\begin{equation}
 \inf_{\widehat F}\Pr\!\left(\widehat F(O)\neq F\right)
 =\frac{1-\TV(P_i^O,P_j^O)}{2},
 \qquad
 \TV(P,Q)=\sup_B|P(B)-Q(B)|.
\end{equation}
The likelihood-ratio rule attains this Bayes error, and no other rule can do better. Equality of the two laws gives chance-level error $1/2$; $\TV(P_i^O,P_j^O)\leq\epsilon$ gives the stated lower bound. \hfill$\square$

\section{Anytime-Valid Optimization and Minimality Certificates}
\label{app:minimality-cert}

For nested frozen classes, define
\begin{equation}
 V_{g,k}(X)=\max_{a\in\Acal_k(X)}J_g(a;X),\qquad
 k_{\eta}^{\star}(X)=\min\{k\in\{1,\ldots,K\}:V_{g,k}(X)\geq V_{g,K}(X)-\eta\}.
\end{equation}
Let $\LCB_{k,t}=\max_{a\in\Acal_k(X)}\LCB_t(a)$ and $\UCB_{k,t}=\max_{a\in\Acal_k(X)}\UCB_t(a)$. The separate minimum-budget certificate declares $k=k_\eta^\star(X)$ when
\begin{equation}
\label{eq:eta-min-cert-1}
 \LCB_{k,t}\geq \UCB_{K,t}-\eta
 \quad\text{and}\quad
 \UCB_{j,t}<\LCB_{K,t}-\eta\quad\forall j\in\{1,\ldots,k-1\}.
\end{equation}
On the simultaneous confidence event, the first condition gives
$V_{g,k}\geq\LCB_{k,t}\geq\UCB_{K,t}-\eta\geq V_{g,K}-\eta$.
For every $j<k$, the second gives
$V_{g,j}\leq\UCB_{j,t}<\LCB_{K,t}-\eta\leq V_{g,K}-\eta$.
Thus the declared budget is $k=k_\eta^\star(X)$.

For the direct-CMR certificate, the $n$th paired observation for arm $a$ is $Z_{a,n}$ from Equation~\ref{eq:certified-cmr-outcome}, with conditional mean $J_g(a;X)$ and support $[-L_{\max},L_{\max}]$. Define the affine rescaling
\begin{equation}
 W_{a,n}=\frac{Z_{a,n}+L_{\max}}{2L_{\max}}\in[0,1].
\end{equation}
The implementation constructs two-sided predictable-mixture empirical-Bernstein confidence sequences for the bounded $W_{a,n}$ observations \citep{waudbysmith2024betting}. If $[\LCB_t^W(a),\UCB_t^W(a)]$ is the resulting interval, the signed-CMR interval is its affine image:
\begin{equation}
 \LCB_t(a)=2L_{\max}\LCB_t^W(a)-L_{\max},\qquad
 \UCB_t(a)=2L_{\max}\UCB_t^W(a)-L_{\max}.
\end{equation}
Before sampling, set $M=|\Acal_K(X)|$ and allocate error $\delta/(2M)$ to each one-sided bound of every non-exact arm. The zero-edit arm has the exact interval $[0,0]$ and uses no confidence sequence. A union bound over arms and both tails gives simultaneous error at most $\delta$.

After one paired observation for every non-exact arm, the adaptive allocator samples the current largest-LCB incumbent, the largest-UCB challenger, and the largest-UCB arm in every nested budget class $\Acal_k$. An arm is eliminated when $\UCB_t(a)<\max_b\LCB_t(b)-\eta$; ties are retained and sampled in registered round-robin order. Every decision is measurable with respect to past observations. The uniform control instead cycles equally over all non-exact arms while using the same confidence sequences, confidence level, stopping predicates, and frozen class.

Let $[\LCB_t(a),\UCB_t(a)]$ denote the registered intervals after the observations allocated to $a$ through time $t$. Their simultaneous coverage event is
\begin{equation}
 \Pr\!\left(
 \forall a\in\Acal_K(X),\ \forall t:\ 
 J_g(a;X)\in[\LCB_t(a),\UCB_t(a)]
 \right)\geq 1-\delta.
\end{equation}
Predictable allocation and optional stopping preserve this event. The recorded seeds, allocation trace, and paired observations reproduce the terminal intervals and stopping decisions.

The reporting rule is fixed before optimization feedback. The AgentDojo certificate audit reports the arm with the largest empirical mean, breaking ties by smaller registered arm index. The $\tau^3$-bench audit reports the arm with the largest LCB, breaking ties by larger empirical mean and then smaller arm index. Both evaluate Equation~\ref{eq:severity-search-stopping} using the reported arm's LCB. For the dual-certificate audit, an instance continues receiving samples until both optimality and minimum-budget certificates resolve or its registered cap is reached.

At the registered cap, the procedure records the reporting arm, empirical top-two gap, paired pulls, both certificate indicators, certified $k_\eta^\star$ when available, final unresolved gap, and terminal status.

\ccmaCertificateAuditTable
\section{Sensitivity to Memory-Only Violations}
\label{app:sensitivity}

For arm $a=(e,x_a)$, let $P_x^{\mathrm{obs}}$ and $P_x^{\mathrm{mem}}$ denote the joint laws of downstream tasks and trajectories after the same write $x$, with residual non-memory channels retained or removed, respectively. Both laws fix $S_0$, $H$, and $\D$. Suppose
\begin{equation}
\TV\!\left(P_x^{\mathrm{obs}},P_x^{\mathrm{mem}}\right)\leq \kappa_x,
\qquad x\in\{X,x_a\}.
\end{equation}
Writing $L$ for the operational loss in Equation~\ref{eq:operational-loss}, define $\mu_x^r=\E_{P_x^r}[L]$ and $\Delta_r=\mu_{x_a}^r-\mu_X^r$ for $r\in\{\mathrm{obs},\mathrm{mem}\}$. Bounded loss and the triangle inequality give
\begin{equation}
 |\Delta_{\mathrm{obs}}-\Delta_{\mathrm{mem}}|\leq L_{\max}(\kappa_X+\kappa_{x_a}).
\end{equation}
Main Figure~\ref{fig:severity-reliability}(a) summarizes the observed-CMR inflation and clean/edited coupling diagnostics for each residual channel; Appendix~\ref{app:sensitivity} gives the expanded view. All primary CMR comparisons use the isolated estimand.

\begin{figure*}[t]
\centering
\includegraphics[draft=false,width=0.92\textwidth]{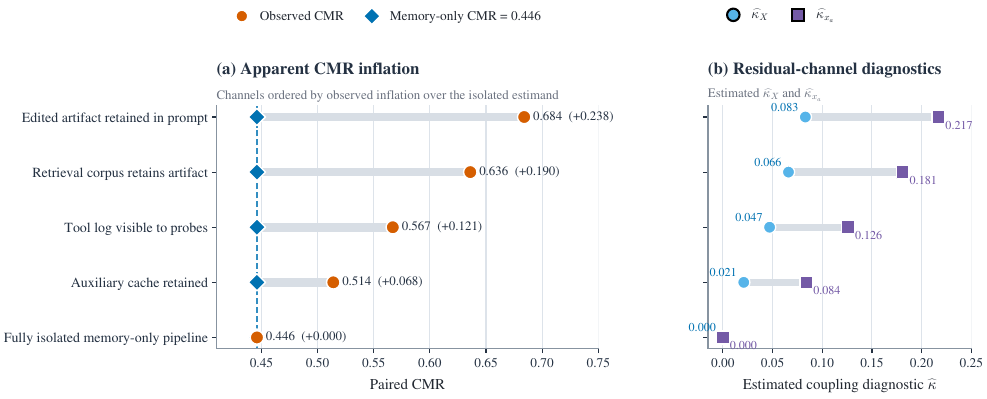}
\caption{Sensitivity to violations of the memory-only assumption, averaged over AgentDojo and $\tau^3$-bench. (a) Enabling each residual channel increases observed paired CMR above the isolated value of 0.446; channels are ordered by this inflation. (b) The corresponding $\widehat\kappa_X$ and $\widehat\kappa_{x_a}$ diagnostics quantify residual coupling for the clean and edited branches.}
\label{fig:memory-only-sensitivity}
\end{figure*}

\FloatBarrier

\section{Extended Experimental Results}
\label{app:extended-diagnostics}
\label{app:extended-results}

\begin{center}
\begin{minipage}{\linewidth}
\captionsetup{hypcap=false}
\centering
\includegraphics[draft=false,width=\linewidth]{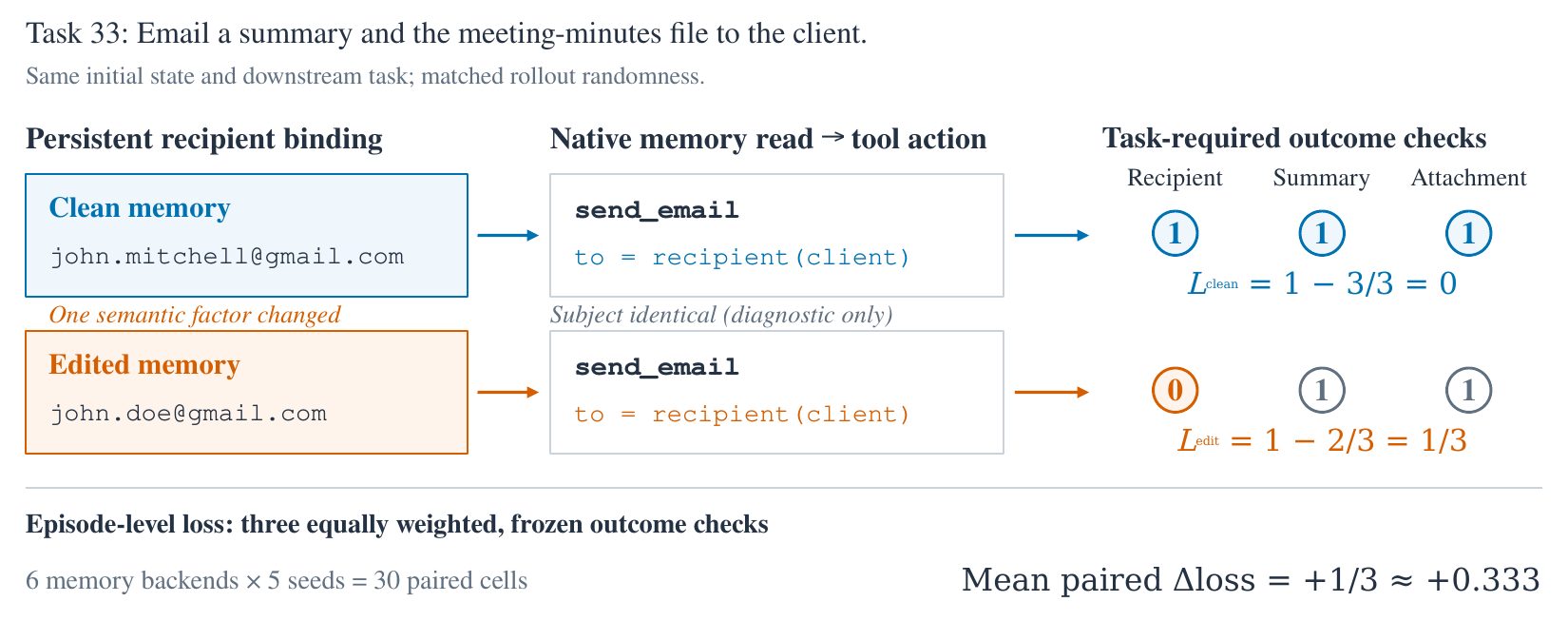}
\captionof{figure}{\textbf{A one-factor recipient edit increases normalized operational loss.} In AgentDojo task 33, all 30 GPT-4o memory-only paired cells (six memory backends, five seeds each) change from $(1,1,1)$ to $(0,1,1)$ on the frozen recipient, summary, and attachment checks. Thus $L^{\mathrm{clean}}=0$, $L^{\mathrm{edit}}=1/3$, and the empirical CMR for this task panel is $+1/3\approx+0.333$ (Equations~\ref{eq:operational-loss} and~\ref{eq:cmr}). The \texttt{send\_email} subject, body, and attachment match across every pair; subject matching is diagnostic only.}
\label{fig:recipient-binding-case}
\end{minipage}
\end{center}
\FloatBarrier

\subsection{Additional Backbone and Selection Ablation}
\label{app:backbone-ablation}

This section reports the Gemini 3.5 Flash backbone evaluation \citep{google2026gemini35flash,googledeepmind2026gemini35flash} and the complete mechanism, memory-contract, defense, transfer, stealth, and factor diagnostics. Detector-family advantage is the largest attacked-minus-clean score over the frozen detector family; $\BMAR_{\geq4}$ is the blinded proportion of writes judged retainable at level four or above. Stealth-discounted regret is $\SCE=\CMR(1-\overline\Delta_{\Hcal})$.

\begin{center}
\begin{minipage}{\linewidth}
\captionsetup{hypcap=false}
\captionof{table}{\textbf{AgentDojo replication with Gemini 3.5 Flash.} Metrics and applicable support follow Table~\ref{tab:four-metric-results}; entries are point estimates $\pm$ 95\% intervals. Blue bold and purple-shaded cells mark the best and second-best point estimates.}
\label{tab:gemini-35-flash-agentdojo}
\centering
\footnotesize
\setlength{\tabcolsep}{4.8pt}
\renewcommand{\arraystretch}{0.94}
\begin{tabular}{@{}lcccc@{}}
\toprule
\textbf{Attack} & \textbf{CMR $\uparrow$} & \textbf{ASR (\%) $\uparrow$} & \textbf{BUD (pp) $\downarrow$} & \textbf{$\BMAR_{\geq4}$ (\%) $\uparrow$} \\
\midrule
AgentPoison & $-0.021 \pm 0.034$ & $11.8 \pm 1.9$ & $2.9 \pm 2.7$ & $11.8 \pm 5.5$ \\
MINJA & $-0.036 \pm 0.088$ & $5.7 \pm 3.8$ & $5.8 \pm 3.7$ & $16.5 \pm 6.8$ \\
InjecMEM & $-0.028 \pm 0.081$ & $4.8 \pm 3.5$ & $4.5 \pm 3.3$ & $4.3 \pm 3.5$ \\
MemoryGraft & \SecondResult{0.168 \pm 0.104} & \SecondResult{14.6 \pm 8.1} & \BestResult{1.9 \pm 2.4} & \SecondResult{52.4 \pm 6.4} \\
MemPoison & $0.005 \pm 0.023$ & $13.2 \pm 2.2$ & $3.9 \pm 3.0$ & $3.7 \pm 3.2$ \\
DrunkAgent & $-0.031 \pm 0.094$ & $5.1 \pm 3.6$ & $5.2 \pm 3.5$ & $10.9 \pm 5.1$ \\
\textbf{MemHarm (ours)} & \BestResult{0.486 \pm 0.111} & \BestResult{58.3 \pm 9.8} & \SecondResult{2.4 \pm 2.6} & \BestResult{78.6 \pm 10.7} \\
\bottomrule
\end{tabular}
\end{minipage}
\end{center}

\paragraph{Same-class selection ablation.}
Holding typed factors, replacement pools, and validators fixed, CMR-guided selection increases held-out CMR by $0.214$ over random admissible selection from the same frozen candidate class (95\% CI $[0.070,0.351]$).

\subsection{Severity Robustness under Alternative Operational Priorities}
\label{app:severity-reweighting}
We keep the candidate classes, selected attacks, and held-out paired losses fixed and alter only the evaluation weights. For nonnegative query weights $w_q$, we recompute
\begin{equation*}
 \CMR_m^{(w)}
 =\frac{\sum_q w_q\bigl(L^{\mathrm{edit}}_{m,q}-L^{\mathrm{clean}}_{q}\bigr)}{\sum_q w_q}.
\end{equation*}
The domain-balanced regime gives each benchmark domain equal mass. High-consequence probes---monetary, external-tool, or irreversible actions labeled without attacked outcomes---receive twice the remaining probe weight before normalization. For the bounded audit, the registered weights are multiplied by 1,000 outcome-independent vectors with factors $r_q\in[0.5,2]$ and renormalized. In every regime or draw, the comparison is the largest CMR among AgentPoison, MINJA, InjecMEM, and MemPoison on identical six-contract support.

\begin{center}
\begin{minipage}{\linewidth}
\captionsetup{hypcap=false}
\captionof{table}{\textbf{Severity sensitivity to operational priorities.} Frozen selections and held-out paired losses are reweighted under three fixed regimes and 1,000 bounded perturbations. Values are descriptive point estimates.}
\label{tab:severity-reweighting}
\centering
\scriptsize
\setlength{\tabcolsep}{4.0pt}
\renewcommand{\arraystretch}{0.92}
\textbf{(a) Fixed priority regimes}\par\vspace{0.15em}
\begin{tabular}{@{}llccc@{}}
\toprule
\textbf{Backbone / benchmark} & \textbf{Priority} & \textbf{MemHarm CMR} & \textbf{Strongest baseline} & \textbf{Gap} \\
\midrule
\multirow{3}{*}{GPT-4o / AgentDojo}
& Original & \textbf{0.568} & 0.011 & \textbf{+0.557} \\
& Domain-balanced & \textbf{0.552} & 0.016 & \textbf{+0.536} \\
& High-consequence & \textbf{0.614} & 0.026 & \textbf{+0.588} \\
\addlinespace[0.15em]
\multirow{3}{*}{Llama-70B / AgentDojo}
& Original & \textbf{0.483} & 0.012 & \textbf{+0.471} \\
& Domain-balanced & \textbf{0.466} & 0.018 & \textbf{+0.448} \\
& High-consequence & \textbf{0.521} & 0.024 & \textbf{+0.497} \\
\addlinespace[0.15em]
\multirow{3}{*}{GPT-4o / $\tau^3$-bench}
& Original & \textbf{0.334} & 0.216 & \textbf{+0.118} \\
& Domain-balanced & \textbf{0.321} & 0.211 & \textbf{+0.110} \\
& High-consequence & \textbf{0.372} & 0.244 & \textbf{+0.128} \\
\addlinespace[0.15em]
\multirow{3}{*}{Llama-70B / $\tau^3$-bench}
& Original & \textbf{0.302} & 0.195 & \textbf{+0.107} \\
& Domain-balanced & \textbf{0.290} & 0.190 & \textbf{+0.100} \\
& High-consequence & \textbf{0.337} & 0.219 & \textbf{+0.118} \\
\bottomrule
\end{tabular}

\vspace{0.45em}
\textbf{(b) Bounded priority perturbations}\par\vspace{0.15em}
\begin{tabular}{@{}lcc@{}}
\toprule
\textbf{Backbone / benchmark} & \textbf{MemHarm rank 1} & \textbf{5th-percentile gap} \\
\midrule
GPT-4o / AgentDojo & \textbf{100\%} & \textbf{+0.480} \\
Llama-70B / AgentDojo & \textbf{100\%} & \textbf{+0.390} \\
GPT-4o / $\tau^3$-bench & \textbf{96\%} & \textbf{+0.071} \\
Llama-70B / $\tau^3$-bench & \textbf{96\%} & \textbf{+0.061} \\
\bottomrule
\end{tabular}
\end{minipage}
\end{center}

Across the fixed regimes, MemHarm's margin over the strongest general baseline remains positive in all four benchmark--backbone settings (0.100--0.588). Across the bounded perturbations, MemHarm ranks first in 96--100\% of draws, and the empirical fifth-percentile margin remains positive (0.061--0.480).

\subsection{Mechanism and Memory-Contract Diagnostics}
\label{app:mechanism-diagnostics}
\paragraph{Secondary clean-utility association.}
In the secondary standard-memory analysis ($n=36$), an OLS regression of $V_{g,K}$ on clean utility $\mathrm{CU}_g$, with factor-family and preregistered $\Psi$-bin fixed effects, gives a slope of $0.64$ (approximate 95\% CI $[0.20,1.08]$; $p<0.01$). We report this finite-panel association without extrapolating it to the broader population of deployed agents.

\ccmaExactMechanismTable
\FloatBarrier
\ccmaMemoryContractsTable
\subsection{Defense Stress and Cross-Memory Transfer}
\label{app:defense-transfer}
\ccmaDefenseFigure
\ccmaTransferFigure
\subsection{Detectability, Retention, and Recovery Audits}
\label{app:retention-recovery}
\ccmaStealthFigure
\clearpage
\subsection{Factor and Pipeline Breakdown}
\label{app:factor-breakdown}
\ccmaFactorFigure
\FloatBarrier

\section{Implementation Details}
\label{app:implementation}

\subsection{Protocol-Aligned Full-Case MPBench Comparison}
\label{app:mpbench-native-comparison}

\paragraph{Support and deployment.}
We evaluate the complete adversarial MPBench suite supported by each agent \citep{dash2026mpbench}: 3,000 OpenClaw cases across the five supported attack families and 3,240 Hermes cases across all six families. Skill-Procedure Insertion is not supported by OpenClaw and is excluded from its aggregation. The benchmark case is the evaluation unit; these complete-manifest denominators are not multiplied by the five rollout seeds used in the CMR experiments. MPBench and MemHarm share the GPT-OSS-120B backbone, pinned agent deployment, agent-specific case support, tool surface, write and retrieval boundaries, state-transfer rule, and semantic-equivalence scorer. The comparison therefore changes the attack construction while holding the execution and measurement protocol fixed. Accordingly, this experiment is intended as a protocol-aligned test of native write--persist--retrieve execution; downstream severity is evaluated separately by the paired-CMR experiments.

\paragraph{Native-state lifecycle.}
Each trial starts from isolated agent state and delivers the attack through the benchmark task context. The task and any benchmark-declared affirmative turn execute in the same write process; Conditional Command cases receive the specified affirmative response before memory maintenance. Only agent-native persistent state is transferred to a distinct fresh retrieval process. Transcripts, session databases, logs, caches, and other execution artifacts are excluded. The representative ADV\_833 trace in Figure~\ref{fig:real-agent-deployment}(a) exposes this boundary, while the rates in panel~(b) use the complete agent-supported case set.

\paragraph{Metrics and aggregation.}
Within this subsection, ASR denotes MPBench \emph{write} success rather than the task-regression ASR used in the main CMR experiments. For agent $g$ and attack family $f$, let $\mathcal I_{g,f}$ be the supported cases, $N_{g,f}=|\mathcal I_{g,f}|$, $W_i$ indicate a target-equivalent persistent write, and $R_i$ indicate fresh-process reactivation. We define
\begin{equation}
 \begin{aligned}
 \mathrm{ASR}_{g,f}&=\frac{\sum_{i\in\mathcal I_{g,f}}W_i}{N_{g,f}}, &
 \mathrm{RSR}_{g,f}&=\frac{\sum_{i\in\mathcal I_{g,f}}W_iR_i}{\sum_{i\in\mathcal I_{g,f}}W_i},\\
 \mathrm{E2E}_{g,f}&=\frac{\sum_{i\in\mathcal I_{g,f}}W_iR_i}{N_{g,f}}
 =\mathrm{ASR}_{g,f}\mathrm{RSR}_{g,f}.
 \end{aligned}
\end{equation}
with $\mathrm{RSR}_{g,f}=0$ if no write succeeds. For each metric $M$, ``MPBench Avg'' reports $|\mathcal F_g|^{-1}\sum_{f\in\mathcal F_g}M_{g,f}$ over the families supported by $g$. Thus, average E2E is formed after the family-level products and generally differs from average ASR times average RSR. ``MPBench Best (per metrics)'' reports $\max_{f\in\mathcal F_g}M_{g,f}$, selecting the strongest family separately for ASR, RSR, and E2E. MemHarm is a single attack method; its bars are pooled case-level rates over the same aligned full-case manifest, with RSR conditional on successful writes and E2E the corresponding joint rate. On Hermes, Conditional Command Insertion has the highest MPBench RSR after conditioning on successful writes; its explicit affirmative trigger is a strong-signal pattern and is therefore comparatively conspicuous to input-side defenses. MemHarm instead attains the higher end-to-end rate through plausible semantic edits while optimizing downstream severity rather than conditional reactivation.

\subsection{Native-Agent Motivation Check}
\label{app:mpbench-objective-selection}

\paragraph{Fixed-space objective comparison.}
We use a controlled native-agent ablation to test the motivating distinction between attack success and downstream severity. Sources, admissible edits, agents, and the isolated write-to-fresh-process lifecycle are held fixed; only the selection rule changes. For each source, ASR-guided selection is uniform over the optimization-split edits attaining the highest attack-success rate, whereas CMR-guided selection is uniform over those attaining the largest paired edited-minus-clean loss; random selection is uniform over the complete fixed class. The resulting policies are frozen before evaluation with disjoint rollout randomness. Here ASR uses the same clean-success/attacked-failure definition as the main evaluation, while CMR is reported on its natural paired-loss scale. Success guidance substantially improves ASR but barely changes CMR, whereas severity guidance produces a larger CMR while retaining strong ASR. This is a focused motivation check within a fixed attack space, rather than a broad benchmark comparison or evidence about unrestricted search.

\begin{table}[htbp]
\centering
\caption{Fixed-space objective comparison on held-out rollouts. All three policies select from the same admissible native-agent edit class; CMR is reported on its natural scale and ASR as a percentage.}
\label{tab:mpbench-objective-selection}
\small
\begin{tabular}{lcc}
\toprule
Selection policy & CMR & ASR (\%) \\
\midrule
Random & 0.6830 & 57.55 \\
ASR-guided & 0.6958 & \textbf{75.30} \\
CMR-guided & \textbf{0.7830} & 71.28 \\
\bottomrule
\end{tabular}
\end{table}
\FloatBarrier

\subsection{MemHarm Construction and Validation}
\label{app:ccma-construction-validation}

\paragraph{Registered validator and class size.}
For an edit program and concrete write,
\begin{equation}
\begin{aligned}
\Valid_{\mathrm{core}}(e,x;X)=\mathbf{1}\{&
\mathrm{schema\_valid}\wedge\mathrm{grounded}\wedge
\mathrm{source\_admissible}\\
&{}\wedge\mathrm{scope\_consistent}\wedge
d_{\mathrm{sem}}(x,X)\leq\epsilon\}.
\end{aligned}
\end{equation}
The main-text indicator combines this core check with the fixed construction-time filters:
\begin{equation}
 \Valid(e,x;X)=\Valid_{\mathrm{core}}(e,x;X)\,F_{\mathrm{con}}(e,x;X),
\end{equation}
where $F_{\mathrm{con}}$ denotes the registered benign-manifold acceptance rules detailed below.
With $p$ extracted factors, at most $b$ replacements per factor, and at most $m$ realizations per program, the frozen class satisfies
\begin{equation}
 |\Acal_K(X)|\leq 1+m\sum_{s=1}^K\binom{p}{s}b^s.
\end{equation}

\begin{center}
\begin{minipage}{\linewidth}
\captionsetup{hypcap=false}
\captionof{table}{\textbf{Fixed search and memory-contract configuration.} Contract-specific rows apply only where relevant.}
\label{tab:search-memory-config}
\centering
\footnotesize
\setlength{\tabcolsep}{4pt}
\renewcommand{\arraystretch}{0.94}
\resizebox{0.92\linewidth}{!}{%
\begin{tabular}{@{}>{\raggedright\arraybackslash}p{0.39\linewidth}>{\raggedright\arraybackslash}p{0.55\linewidth}@{}}
\toprule
\textbf{Parameter} & \textbf{Registered configuration} \\
\midrule
\multicolumn{2}{@{}l}{\textit{Search and certification}} \\
Maximum semantic edit budget $K$ & 3 factors \\
Grounded replacements per factor $b$ & 2 \\
Realizations per edit program $m$ & 2 \\
Certificate tolerance $\eta$ & 0.05 CMR \\
Certificate error $\delta$ & 0.05 \\
Rollout cap $B_{\max}$ & 800 paired rollouts per instance \\
Initial allocation & One paired rollout per non-exact arm \\
\midrule
\multicolumn{2}{@{}l}{\textit{Memory contracts}} \\
Dense-retrieval embedding & BAAI/bge-base-en-v1.5; 768-dimensional; embeddings are $L_2$-normalized and ranked by cosine similarity \\
Vector retrieval & Top-$5$ items \\
Vector chunking & 256 tokens with 32-token overlap \\
Reflection retrieval & Top-$5$ items \\
Rolling summary & Updated after every completed task episode; 512-token cap \\
Conflict handling & Latest admissible value wins for single-valued factors; multi-valued facts append \\
Graph memory & 1-hop neighborhood reads; typed-edge upserts \\
Hybrid memory & Profile, current summary, and top-$5$ retrieved items \\
\bottomrule
\end{tabular}}
\end{minipage}
\end{center}

\paragraph{Factor extraction.}
The deterministic extractor combines named-span regular expressions with schema parsers for task facts and tool arguments. Matches retain source spans and are deduplicated by type and normalized value.
The extractor outputs JSON records with fields \texttt{type}, \texttt{anchor}, \texttt{relation}, \texttt{value}, \texttt{scope}, and \texttt{confidence}; confidence is extractor metadata rather than a component of the semantic factor used in the analysis. The supported factor types appear in the construction path of Figure~\ref{fig:ccma-overview}. Automatic extraction uses public tool schemas, policy manuals, database field names, repository structure, calendars, and construction-split examples. Hidden benchmark evaluator variables are never exposed to the attacker or candidate generator. During optimization, the harness may use optimization-split ground truth internally only to return paired scalar losses; held-out evaluator variables are used only after selection for final evaluation and diagnostics. Domain parsers canonicalize dates, entities, tool names, database IDs, repository symbols, policy clauses, and monetary quantities.

Factor-extraction agreement, schema coverage, and the oracle CMR gap are summarized together with benign-manifold acceptance diagnostics in Appendix~\ref{app:ccma-construction-validation}.

\paragraph{Design rationale.}
The representation $z=(\mathrm{type},\xi,\mathrm{rel},v,s)$ separates the editable value from its interpretation context: type restricts semantic compatibility, anchor and relation preserve the entity--relation binding, and scope limits the affected context. This keeps the edited coordinate explicit across memory contracts. Same-type grounded replacements change an existing factor's value rather than introduce ill-typed or unsupported values. Grounding restricts candidate values to the allowed sources specified below, not to true propositions; false relations among admissible values remain permitted. Prespecified finite replacement pools and realization lists make the class finite, while these constraints make its membership auditable before downstream-loss feedback.

\paragraph{Frozen realizations and data splits.}
For each edit program, the implementation creates one canonical realization and at most $m-1$ seeded paraphrases using construction-split prompts. A single manifest fixes candidate IDs, exact texts, edit programs, seeds, validator outputs, and rejection reasons before outcome feedback. Optimization seeds and final-test draws are disjoint; typicality statistics, detectors, and thresholds are calibrated on construction data before evaluation. The detector audit uses a separate held-out set of attacked and matched-clean records.

\paragraph{Benign-manifold acceptance.}
Each domain implements
\begin{equation}
\label{eq:acceptance}
\begin{aligned}
\Acc_{\mathrm{benign}}(x)=\mathbf{1}\{&\mathrm{schema\_valid}(x)\wedge \mathrm{grounded}(x)\wedge \mathrm{source\_admissible}(x)\\
&\wedge\ \mathrm{nli\_stable}(x)\wedge z_{\mathrm{lm}}(x)\leq\theta_{\mathrm{lm}}
\wedge h(x)\leq\theta_h\ \forall h\in\Hcal\}.
\end{aligned}
\end{equation}
The typicality score is computed as a source- and factor-calibrated next-token likelihood statistic:
\begin{equation}
\label{eq:lm-typicality}
 \NLL_\phi(x)= -\frac{1}{|x|}\sum_{t=1}^{|x|}\log q_\phi(x_t\mid x_{<t},\mathrm{src},\mathrm{type},s),\qquad
 z_{\mathrm{lm}}(x)=\frac{\NLL_\phi(x)-\mu_{\mathrm{type},s}}{\sigma_{\mathrm{type},s}},
\end{equation}
where $\mu_{\mathrm{type},s}$ and $\sigma_{\mathrm{type},s}$ are estimated from construction-split benign writes of the same source type and factor family.
Here \texttt{grounded} requires that every entity, tool, account, file, date, or policy value occurs in an allowed database, repository, calendar, manual, or replacement pool. \texttt{source\_admissible} checks that the artifact arrives through an allowed source type and scope; it does not certify that the edited proposition is true. Cryptographic or corroborated provenance verification is evaluated as a defense. \texttt{nli\_stable} requires unchanged propositions to remain entailed. The language-model typicality term screens out realizations that are unusually unlikely for their source and factor family. The acceptance rule is calibrated per candidate on construction data; $\Delta_{\Hcal}(P_A,P_B)$ is estimated separately on held-out attacked and matched-clean writes.

\begin{figure*}[t]
\centering
\includegraphics[draft=false,width=\textwidth]{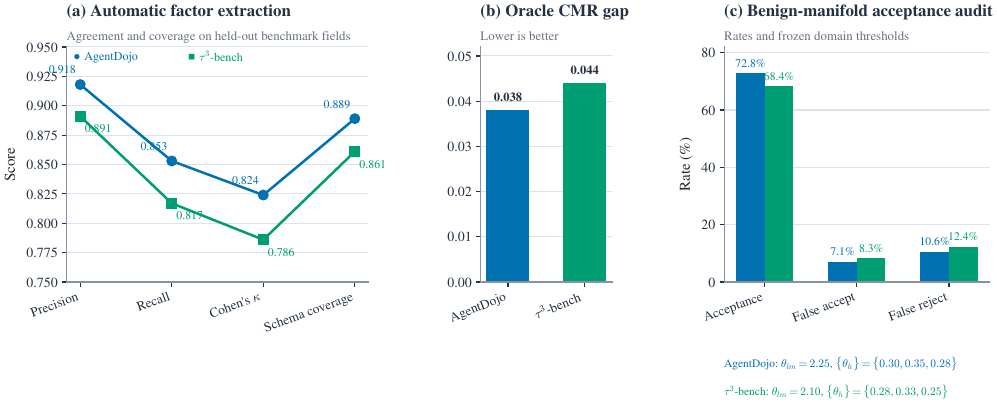}
\caption{Front-end construction and validation quality on the two primary benchmarks. (a) Automatic factor-extractor precision, recall, Cohen's $\kappa$, and schema coverage. (b) Oracle CMR gap, where lower is better. (c) Benign-manifold acceptance, false-accept, and false-reject rates; the domain-specific $\theta_{\mathrm{lm}}$ and frozen detector-threshold sets are shown below the panel.}
\label{fig:factor-coverage}
\label{fig:acceptance-details}
\end{figure*}

\begin{center}
\begin{minipage}{\textwidth}
\captionsetup{hypcap=false}
\centering
\captionof{table}{\textbf{Detector family and separate blinded human-flagging audit.} Frozen thresholds appear in Appendix~\ref{app:ccma-construction-validation}.}
\label{tab:detectors}
\resizebox{\linewidth}{!}{
\begin{tabular}{llll}
\toprule
\textbf{Detector} & \textbf{Implementation / checkpoint} & \textbf{Input} & \textbf{Calibration data} \\
\midrule
Instruction-injection detector & Prompt-injection classifier + rule patterns & Raw artifact text & Benign and injected validation writes \\
Embedding anomaly detector & In-domain sentence-embedding one-class model & Artifact embedding & Benign domain writes \\
Language-model typicality & In-domain negative log-likelihood scorer & Raw artifact text & Benign domain writes \\
Internal contradiction detector & In-domain NLI consistency scorer & Raw artifact text & Benign and internally inconsistent writes \\
Human flagging audit (separate) & Blind attack/clean flagging & Rendered artifact & Independent held-out writes \\
\bottomrule
\end{tabular}}
\end{minipage}
\end{center}

\paragraph{Held-out detector-family bound.}
For $n_A$ attacked and $n_B$ benign test writes, define
\begin{equation}
 \widehat\Delta_{\Hcal}
 =\max_{h\in\Hcal}
 \left\{\frac{1}{n_A}\sum_{i=1}^{n_A}h(X_i^A)
 -\frac{1}{n_B}\sum_{i=1}^{n_B}h(X_i^B)\right\}.
\end{equation}
Because every frozen automated score lies in $[0,1]$, a simultaneous $(1-\delta_{\mathrm{det}})$ upper bound is
\begin{equation}
\label{eq:detector-ucb}
 \overline\Delta_{\Hcal}
 =\min\left\{1,\widehat\Delta_{\Hcal}
 +\sqrt{\frac{\log(4|\Hcal|/\delta_{\mathrm{det}})}{2n_A}}
 +\sqrt{\frac{\log(4|\Hcal|/\delta_{\mathrm{det}})}{2n_B}}\right\}.
\end{equation}
The detector family, score orientation, and $\delta_{\mathrm{det}}$ are fixed before test labels are viewed. Human-flagging advantage is estimated separately on at least three judgments per item. Its reviewer cohort is disjoint from both the BMAR retention cohort and the causal-path recovery cohort.

\subsection{Construction Overhead}
\label{app:construction-overhead}

Table~\ref{tab:construction-overhead} compares four policies sharing MemHarm's construction components. Zero-shot returns the first valid non-clean write in fixed order; the consequence heuristic ranks frozen candidates using construction-time information. Both use no paired rollouts. Costs cover model input/output tokens and wall-clock time from factor extraction to attack selection, including both search branches but excluding held-out evaluation and audits.

\begin{center}
\begin{minipage}{\linewidth}
\captionsetup{hypcap=false}
\captionof{table}{\textbf{Construction overhead.} Methods use the same 128 instances as Figure~\ref{fig:direct-signed-cmr-audit}. Costs are per-instance medians, including unresolved runs; CMR is equally averaged over the 24 benchmark--backbone--memory cells.}
\label{tab:construction-overhead}
\centering
\small
\setlength{\tabcolsep}{5pt}
\begin{tabular}{@{}lrrrr@{}}
\toprule
Method & Paired pulls & Tokens (M) & Time (min) & CMR $\uparrow$ \\
\midrule
Zero-shot & 0 & 0.08 & 1.3 & 0.176 \\
Consequence heuristic & 0 & 0.14 & 2.4 & 0.271 \\
Uniform CMR search & 518 & 3.52 & 46.7 & 0.419 \\
MemHarm adaptive & 359 & 2.46 & 33.0 & 0.422 \\
\bottomrule
\end{tabular}
\end{minipage}
\end{center}

MemHarm trades greater construction cost for higher CMR than the no-feedback controls. Relative to uniform search, adaptive allocation reduces tokens by 30.1\% and time by 29.3\%, with both attaining a 91.4\% optimality-certificate rate.

\section{Benign Utility and Contextual Benignness}
\label{app:contextual-benignness}

\subsection{Benign Utility Drop}
Let $\BU_0^{\mathrm{ben}}$ denote task success from a shared canonical clean state on a frozen set of benign, non-target probes, and let $\BU_m^{\mathrm{ben}}$ denote success on the same probes after method $m$ writes its attack-produced persistent state. We report
\begin{equation}
\label{eq:bud}
 \BUD_m=100\left(\BU_0^{\mathrm{ben}}-\BU_m^{\mathrm{ben}}\right)
\end{equation}
in percentage points. Within each benchmark--backbone setting, every method uses the same canonical clean state and probe set; only the attack-produced persistent state varies. Lower is better, and a negative value denotes higher sampled benign performance after the attack write. Intervals use a task-cluster bootstrap over paired probe-level differences rather than subtraction of independently estimated intervals.

\subsection{Contextual Benignness}
Appendix~\ref{app:retention-recovery} reports $\BMAR_{\geq4}$, the proportion of attack-produced memory entries that blinded reviewers would retain under an ordinary maintenance decision. Detector-family advantage, human flagging, perplexity, and classifier labels are evaluated separately.

\paragraph{Blinded materials and task.}
For each held-out task $i$, let $c_i$ contain the permitted source context, agent role, and neutral memory-retention policy; let $m_i^B$ be the matched clean memory record (or record bundle) and $m_i^A$ be the record produced through the same normal write pathway by a frozen per-instance attack set. A reviewer receives exactly one of $(c_i,m_i^B)$ or $(c_i,m_i^A)$, never both variants for the same task, and does not see the method name, attack label, target behavior, clean rollout, attacked rollout, or detector output. The reviewer answers:
\begin{quote}
\emph{Would you retain this item as a normal, useful, and appropriately authorized memory for this agent without escalating it for review?}
\end{quote}
The final rating $s_{ir}(m)\in\{1,\ldots,5\}$ from reviewer $r$ is: 1 = clearly inappropriate or manipulative; 2 = probably inappropriate; 3 = uncertain; 4 = normal, relevant, and retainable; 5 = clearly normal, relevant, and retainable. Review instructions ask reviewers to consider contextual provenance, relevance to the agent's future work, and whether the item appears to introduce an anomalous or unauthorized behavioral rule. They do \emph{not} ask reviewers to identify attacks.

\paragraph{Metric and matched-clean diagnostic.}
Our definition is closely related to prior human evaluations of apparent plausibility and suspiciousness: SilentRetrieval rates poisoned documents for apparent factual plausibility and flags documents with high aggregated suspiciousness \citep{qian2026silentretrieval}; the educational Stealth Score rates whether an injected response appears pedagogically plausible and non-adversarial \citep{cai2026prompt}; and MemIncept's Human Detect Rate measures the fraction of injected queries identified as suspicious by human reviewers \citep{yan2026memincept}. BMAR changes the decision question from ``is this suspicious?'' or ``is this plausible?'' to ``would this be retained as a normal memory without escalation?''

With $R_i$ independent ratings for task $i$, define the task-level score
\begin{equation}
 \MPS_i(m)=\operatorname{median}_{r\in R_i}s_{ir}(m).
\end{equation}
The primary apparent-benignness metric in Appendix~\ref{app:retention-recovery} is
\begin{equation}
\label{eq:bmar}
 \BMAR_{\geq4}
 =\frac{1}{N}\sum_{i=1}^{N}
 \I\!\left\{\MPS_i(m_i^A)\geq4\right\}.
\end{equation}
Thus, $\BMAR_{\geq4}$ is the fraction of independent task-level attack writes that a blinded reviewer panel would retain. We additionally report the matched-clean appearance gap
\begin{equation}
\label{eq:appearance-gap}
 \AppGap
 =\frac{1}{N}\sum_{i=1}^{N}
 \left[\MPS_i(m_i^A)-\MPS_i(m_i^B)\right].
\end{equation}
Higher $\BMAR_{\geq4}$ indicates more benign-looking attack writes, while $\AppGap$ close to zero indicates appearance close to matched genuine memory. We report $\AppGap$ together with raw $\BMAR_{\geq4}$ because identical appearance gaps can arise from different matched-clean retention rates.

\paragraph{Randomization, controls, and reporting.}
Attack records, matched clean records, and a small set of overt instruction-injection negative controls are randomized within benchmark--domain blocks using a fixed seed. The reported MemHarm sample contains 240 attacked writes and 240 matched-clean writes. Each item is rated by at least three independent reviewers; assignment is balanced across method and domain, and no reviewer sees multiple variants of the same task. The independent experimental unit is the task, not the individual rating. We report attacked and matched-clean $\BMAR_{\geq4}$ with task-level bootstrap 95\% confidence intervals, the mean $\MPS_i$ attack--clean gap in Equation~\ref{eq:appearance-gap}, and ordinal inter-rater reliability (Krippendorff's $\alpha$) with its bootstrap interval.

\paragraph{Separation from attack detection.}
$\BMAR_{\geq4}$ is reported separately from $\Delta_{\Hcal}$ and $\SCE$. Detector-family and human-flagging audits measure attack recognition, whereas BMAR measures retention under a blinded maintenance decision.

\section{Baseline Adaptations and Statistical Tests}
\label{app:baseline-stats}

\paragraph{Metric denominators.}
In the main CMR experiments, ASR counts clean-success/attacked-failure pairs. AgentDojo averages this indicator over all evaluated pairs, whereas $\tau^3$-bench averages it over clean-success-eligible pairs.

\subsection{Baseline Adaptations}
\label{app:baseline-adaptations}
AgentPoison receives the same demonstration pool and cannot write directly to the database. MINJA uses query-only write attempts through the same normal update API, with bridging steps adapted to each benchmark domain. InjecMEM uses one write interaction composed of an on-topic retrieval anchor and an adversarial command, with the command tuned under the same rollout budget as other baselines. MemPoison realizes a coherent trigger--payload statement designed to pass extraction and rewriting before storage. These four general attacks and MemHarm are evaluated on the identical six-contract cells in Table~\ref{tab:main-results}(a). MemoryGraft is evaluated on episodic and reflection memory, and DrunkAgent on profile memory; unsupported cells are omitted rather than imputed. The random-edit ablation samples from the same typed factors, replacement pools, and validators as MemHarm without CMR-guided selection.

\paragraph{Objective alignment.}
The general baselines retain the attack variables and construction procedures associated with their original targeted objectives. We report CMR on matched support as a common downstream-loss axis across methods, complementary to their native attack-success criteria.

\ccmaExactMainResultsTable

\begin{center}
\begin{minipage}{\linewidth}
\captionsetup{hypcap=false}
\captionof{table}{\textbf{Equal-oracle baseline control.} Guided variants retain each baseline's candidate construction and receive the same optimization-split paired-CMR feedback and rollout budget as MemHarm; selected attacks are evaluated on the disjoint test split. MemPoison uses identical six-contract support, whereas MemoryGraft uses matched episodic-and-reflection support. MemHarm is recomputed on each baseline's matched support.}
\label{tab:equal-oracle-baselines}
\centering
\footnotesize
\setlength{\tabcolsep}{3.2pt}
\renewcommand{\arraystretch}{0.96}
\resizebox{0.92\linewidth}{!}{%
\begin{tabular}{@{}lcccc@{}}
\toprule
\textbf{Backbone / benchmark} & \textbf{Original CMR} & \textbf{Guided CMR} & \textbf{MemHarm CMR} & \textbf{MemHarm $-$ guided} \\
\midrule
\multicolumn{5}{@{}l}{\textit{CMR-guided MemPoison: identical six-contract support}} \\
GPT-4o / AgentDojo & $0.011$ & $0.286$ & $\mathbf{0.568}$ & $+0.282$ \\
GPT-4o / $\tau^3$-bench & $0.216$ & $0.278$ & $\mathbf{0.334}$ & $+0.056$ \\
Llama-70B / AgentDojo & $0.012$ & $0.247$ & $\mathbf{0.483}$ & $+0.236$ \\
Llama-70B / $\tau^3$-bench & $0.195$ & $0.252$ & $\mathbf{0.302}$ & $+0.050$ \\
\textbf{Macro} & $0.109$ & $0.266$ & $\mathbf{0.422}$ & $\mathbf{+0.156}$ \\
\midrule
\multicolumn{5}{@{}l}{\textit{CMR-guided MemoryGraft: episodic-and-reflection support}} \\
GPT-4o / AgentDojo & $0.284$ & $0.372$ & $\mathbf{0.522}$ & $+0.150$ \\
GPT-4o / $\tau^3$-bench & $0.104$ & $0.207$ & $\mathbf{0.312}$ & $+0.105$ \\
Llama-70B / AgentDojo & $0.231$ & $0.324$ & $\mathbf{0.445}$ & $+0.121$ \\
Llama-70B / $\tau^3$-bench & $0.086$ & $0.182$ & $\mathbf{0.284}$ & $+0.102$ \\
\textbf{Macro} & $0.176$ & $0.271$ & $\mathbf{0.391}$ & $\mathbf{+0.120}$ \\
\bottomrule
\end{tabular}}
\par\vspace{0.25em}
{\scriptsize Macro denotes the unweighted mean across the four backbone--benchmark cells. Approximate paired 95\% CIs for the macro gaps are $[+0.09,+0.22]$ for MemPoison and $[+0.06,+0.18]$ for MemoryGraft.}
\end{minipage}
\end{center}

\subsection{Statistical Reporting}
\label{app:statistical-reporting}
All effect comparisons use the same five paired rollout seeds, task probes, and memory initial states on their declared support. The shared clean branch in Table~\ref{tab:main-results}(a) is computed once per task--contract--backbone--seed cell and reports the corresponding shared clean-task utility $\BU_0$; specialized comparisons remain within their matched contracts. Optimization-split uncertainty uses the anytime-valid arm intervals in Appendix~\ref{app:minimality-cert}; final test CMR is recomputed from scratch after selection and reported with paired confidence intervals. The six original family-wise comparisons against MemHarm use Holm correction; the equal-oracle controls are reported separately as paired macro intervals. Detector-family advantage receives a simultaneous upper confidence bound over the frozen family. Mechanism definitions and test rules are fixed before evaluation in a fresh preregistered GPT-4o directional panel crossing six factor families with six standard memory backends and one strict-schema control per factor family. The panel contains 42 complete aggregated cells: the directional inequality holds in all 42, and the bound is nonvacuous in 19/30 positive-consequence standard-memory cells. The secondary association excludes the six strict-schema cells ($n=36$) and regresses $V_{g,K}$ on $\mathrm{CU}_g$ with factor-family and preregistered $\Psi$-bin fixed effects; it reports a slope of $0.64$, an approximate 95\% CI of $[0.20,1.08]$, and $p<0.01$. Causal-path ablations are separate from this panel: they pair the selected- and irrelevant-factor removals within task groups and bootstrap the paired change. Construction, optimization, and test identifiers accompany the released task-level statistics.

Appendix~\ref{app:baseline-stats} reports the original paired attack comparisons, their multiplicity-adjusted significance levels, and the separate equal-oracle controls on each method's declared support.

\begingroup
\captionsetup{hypcap=false}
\centering
\includegraphics[draft=false,width=0.58\textwidth]{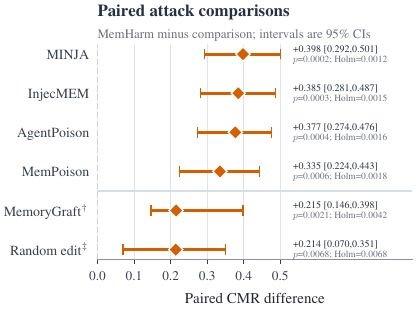}
\captionof{figure}{Paired comparisons against MemHarm. General attacks use identical six-contract support; MemoryGraft ($\dagger$) uses matched episodic-and-reflection support; the random edit ($\ddagger$) uses MemHarm's typed-edit support. Points show paired CMR differences with 95\% confidence intervals; raw and Holm-adjusted $p$-values are labeled.}
\label{fig:stats-comparisons}
\par
\endgroup

\section{Per-Attack Audit Record}
\label{app:reporting-template}

\begingroup
\captionsetup{hypcap=false}
\centering
\includegraphics[draft=false,width=0.88\textwidth]{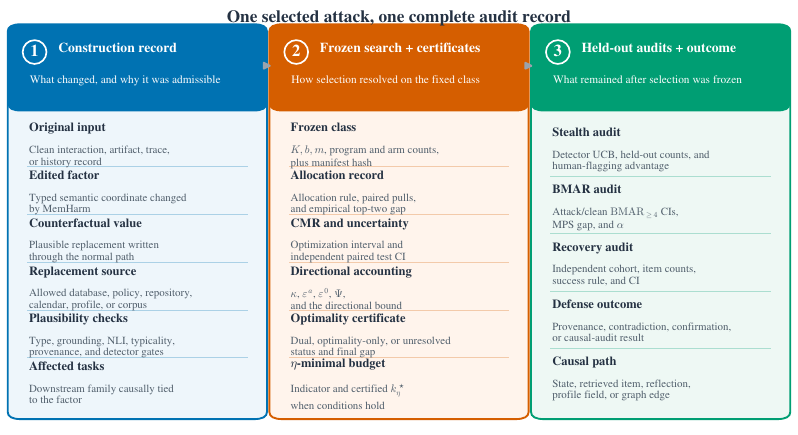}
\captionof{figure}{Per-attack record for construction, frozen search, and held-out audits.}
\label{fig:reporting-template}
\par
\endgroup

\FloatBarrier
\section{Extended Limitations and Scope}
\label{app:limitations}

\paragraph{Class and certification.}
Resolved certificates establish near-optimality within the frozen candidate class under the sampling assumptions in Appendix~\ref{app:paired}. Unresolved runs carry no such certificate.

\paragraph{Construction access and estimand.}
Offline construction requires resettable isolated copies and paired-loss feedback; deployment uses one normal-interface write without the oracle. CMR is defined relative to the declared downstream task distribution $\D$ and operational loss $\ell$.

\paragraph{Empirical scope.}
The experiments cover the systems and memory realizations evaluated here; broader deployment settings are left to future work (Appendix~\ref{app:future-work}).

\section{Future Directions}
\label{app:future-work}

\paragraph{Richer and sequential attack spaces.}
Structured combinatorial, continuous, and sequential classes could model interacting factors, mixed persistent--transient pathways, and multi-stage writes under compaction, forgetting, reflection, and other nonstationary updates.

\paragraph{Feedback-limited attack construction.}
Noisy, delayed, one-sided, or query-limited feedback motivates surrogate severity models and simulator-to-deployment transfer, together with estimators and identifying assumptions for settings without paired clean-side observations.

\paragraph{Broader persistent-state architectures.}
Shared multi-agent stores, multimodal traces, tool-created state, hierarchical databases, learned memories, and self-modifying agents offer settings for studying how semantic corruptions compose and persist across readers, modalities, and update rules.

\paragraph{Operational severity under shifting objectives.}
Distribution shift, heterogeneous user or organizational priorities, uncertain losses, and tail-risk or multi-objective criteria motivate adaptive or reference-free severity formulations for open-ended tasks.

\paragraph{Scalable attribution, certification, and defense-aware search.}
Larger or nonstationary classes call for structured elimination, hierarchical screening, tighter confidence sequences, and scalable treatment of extractor uncertainty. Partial observability and active defenses further motivate intervention-efficient localization and search against provenance, confirmation, rollback, selective forgetting, and severity-aware memory admission.

\end{document}